\documentclass[nohyperref]{article}

\usepackage{microtype}
\usepackage{graphicx}
\usepackage{booktabs} %

\usepackage{hyperref}

\usepackage[accepted]{icml2022}

\usepackage{amsmath}
\usepackage{amssymb}
\usepackage{mathtools}
\usepackage{amsthm}

\usepackage[textsize=tiny]{todonotes}

\usepackage{bbm}
\usepackage{booktabs}
\usepackage{enumitem}
\usepackage{multirow}
\usepackage{graphicx}
\usepackage{tikz}
\usepackage[font=small]{caption}
\usepackage{tikz}
\usetikzlibrary{positioning,shapes,fit,calc,arrows.meta, patterns}
\usepackage{subcaption}
\usepackage{wrapfig}

\newcommand{\e}{\epsilon}
\newcommand{\y}{\gamma}

\newcommand{\ta}{\theta}

\newcommand{\E}{\mathbb{E}}
\newcommand{\N}{\mathcal{N}}
\newcommand{\lb}{\left [}
\newcommand{\rb}{\right ]}
\newcommand{\lp}{\left (}
\newcommand{\rp}{\right )}

\newcommand{\Loss}{\mathcal{L}}

\newtheorem{theorem}{Theorem}
\newtheorem{corollary}{Corollary}
\newtheorem{observation}{Observation}
\newtheorem{proposition}{Proposition}

\usepackage{varioref}
\labelformat{equation}{(#1)}

\makeatletter
\newcommand{\setword}[2]{%
  \phantomsection
  #1\def\@currentlabel{\unexpanded{#1}}\label{#2}%
}
\makeatother

\newcommand{\cblock}[1]{
 \begin{tikzpicture}
   [
   node/.style={square, minimum size=10mm, thick, line width=0pt},
   ]
   \node[fill={#1}] {};
 \end{tikzpicture}%
}

\newcommand{\cblockhtml}[1]{
 \definecolor{#1}{HTML}{#1}
 \begin{tikzpicture}
   [
   node/.style={square, minimum size=10mm, thick, line width=0pt},
   ]
   \node[fill={#1}] {};
 \end{tikzpicture}%
}

\newcommand{\ctriangle}[1]{
\begin{tikzpicture}
\draw[fill={#1}, draw={#1}] (0.1,0.1) -- (-0.1,0.1) -- (0,-0.1) -- cycle;
\end{tikzpicture}
}

\newcommand{\cinvtriangle}[1]{
\begin{tikzpicture}
\draw[fill={#1}, draw={#1}] (-0.1,-0.1) -- (0.1,-0.1) -- (0,0.1) -- cycle;
\end{tikzpicture}
}

\newcommand{\ccircle}[1]{
\begin{tikzpicture}
\draw[fill={#1}, draw=#1] (0,0) circle (3pt);
\end{tikzpicture}
}

\definecolor{sb_blue}{RGB}{31, 119, 180}
\definecolor{sb_orange}{RGB}{255, 127, 14}
\definecolor{sb_green}{RGB}{44, 160, 44}
\definecolor{sb_red}{RGB}{214, 39, 40}
\definecolor{sb_purple}{RGB}{148, 103, 189}

\definecolor{my_blue}{RGB}{70, 119, 208}

\definecolor{my_cold_1}{HTML}{a0da39}
\definecolor{my_cold_2}{HTML}{4ac16d}
\definecolor{my_cold_3}{HTML}{1fa187}
\definecolor{my_cold_4}{HTML}{277f8e}
\definecolor{my_cold_5}{HTML}{365c8d}
\definecolor{my_cold_6}{HTML}{46327e}

\definecolor{my_warm_1}{HTML}{ffaa5c}
\definecolor{my_warm_2}{HTML}{f8765c}
\definecolor{my_warm_3}{HTML}{d3436e}
\definecolor{my_warm_4}{HTML}{982d80}
\definecolor{my_warm_5}{HTML}{5f187f}
\definecolor{my_warm_6}{HTML}{221150}

\icmltitlerunning{The Bellman Error is a Poor Replacement for Value Error}

\begin{document}

\twocolumn[
\icmltitle{Why Should I Trust You, Bellman? \\The Bellman Error is a Poor Replacement for Value Error}

\icmlsetsymbol{equal}{*}

\begin{icmlauthorlist}
\icmlauthor{Scott Fujimoto}{mila,google}
\icmlauthor{David Meger}{mila}
\icmlauthor{Doina Precup}{mila,deepmind}
\icmlauthor{Ofir Nachum}{google}
\icmlauthor{Shixiang Shane Gu}{google}
\end{icmlauthorlist}

\icmlaffiliation{mila}{Mila, McGill University}
\icmlaffiliation{google}{Google Research, Brain Team}
\icmlaffiliation{deepmind}{DeepMind}
\icmlcorrespondingauthor{Scott Fujimoto}{scott.fujimoto@mail.mcgill.ca}

\icmlkeywords{Machine Learning, ICML, Reinforcement Learning, Bellman, Policy Evaluation, Off-Policy}

\vskip 0.3in
]

\printAffiliationsAndNotice{}  %

\begin{abstract}
In this work, we study the use of the Bellman equation as a surrogate objective for value prediction accuracy. While the Bellman equation is uniquely solved by the true value function over all state-action pairs, we find that the Bellman error (the difference between both sides of the equation) is a poor proxy for the accuracy of the value function. In particular, we show that (1) due to cancellations from both sides of the Bellman equation, the magnitude of the Bellman error is only weakly related to the distance to the true value function, even when considering all state-action pairs, and (2) in the finite data regime, the Bellman equation can be satisfied exactly by infinitely many suboptimal solutions. This means that the Bellman error can be minimized without improving the accuracy of the value function. We demonstrate these phenomena through a series of propositions, illustrative toy examples, and empirical analysis in standard benchmark domains.
\end{abstract}

\section{Introduction}

In reinforcement learning (RL), value functions are a measure of performance of a target policy. Value functions are an important quantity in RL as they can be used to inform decision-making. 
Consequently, many modern RL algorithms rely on a value function in some capacity~\citep{DQN, gu2016continuous, ppo, fujimoto2018addressing, badia2020agent57}.

The Bellman equation is a fundamental relationship in RL which relates the value of a state-action pair to the state-action pair that follows and is uniquely satisfied over all state-action pairs by the true value function.
The existence of the Bellman equation suggests a straightforward approach to approximate value function learning, where a function is trained to minimize the Bellman error (the difference of both sides of the equation). The Bellman equation has played a prominent role in many historically significant approaches~\citep{schweitzer1985generalized, baird1995residual, bradtke1996linear, antos2008learning, sutton2009fast}, as well as the more modern family of deep RL algorithms~\citep{DQN, DDPG, gu2016continuous, hessel2018rainbow}. 

This work aims to better understand the relationship between the Bellman error and the accuracy of value functions through theoretical analysis and empirical study. 
We do so by focusing on the setting of policy evaluation, which presents the task of learning the value function of a target policy with data gathered from a possibly separate and unknown behavior policy. Policy evaluation, which is a subcomponent %
of many RL algorithms, is an ideal setting for evaluating value functions as it provides a clear metric of performance (value error: the distance to the true value function) and provides consistency across trials (via a fixed dataset and policy). Our key discoveries are as follows: 

\textbf{The Bellman error is a poor replacement for value error.} We find that given an arbitrary value function, low Bellman error does not indicate low value error. 
This problem is highlighted by experiments which show that value functions trained to minimize the Bellman error directly~\citep{baird1995residual} have lower average Bellman error, but higher average value error, than value functions trained by iterative methods~\citep{ernst2005tree}. This inverse relationship 
holds even when evaluated over on-policy data~(\autoref{fig:both}), and only worsens further with off-policy data~(\autoref{table:OPE_obj}). 
We show that this breakdown between Bellman error and value error is due to the following two phenomena.

\textbf{The magnitude of the Bellman error hides bias.} The Bellman error of a value function is the difference between the value estimates on each side of the Bellman equation. %
If the value error of these estimates share the same sign (i.e.\ they are biased in the same direction), %
then they can cancel each other out. As a result, biased value functions will have lower Bellman error than unbiased value functions, even if they have the same absolute value error. %

\textbf{Missing transitions breaks the Bellman equation.} The Bellman equation is meant to consider the entire MDP and all state-action pairs. We show that when the Bellman equation is instead evaluated over an incomplete dataset, it can be satisfied exactly by infinitely many suboptimal solutions. It follows that minimizing the Bellman error over a finite dataset is not guaranteed to improve the accuracy of a value function. %
In our practical experiments, we show that when trained with off-policy data, methods which directly reduce the Bellman error find value functions 
that have near zero Bellman error, but large value error, for any state-action pair in the dataset.

We demonstrate the existence of these phenomena through theoretical analysis, toy examples (\autoref{sec:BE}), and experimentation in standard benchmarks (\autoref{sec:experiments}). We find that these problems extend beyond carefully designed counterexamples, as they are consistent across domains, and can appear in a clear, and often extreme, manner while using common algorithms with standard hyperparameters. 

Our work highlights problems with using the Bellman error as a signal or objective, particularly in the off-policy setting. We aim to provide a better understanding of performance gaps in RL methods: \textit{``why does directly minimizing the Bellman error perform poorly?''} and Bellman equation-based loss functions: \textit{``why does the loss function not correspond with performance?''} 
Our findings point to an underappreciation of the importance of finite data in widely used objectives and we encourage the community to place a higher emphasis on practical settings. %

\section{Background}\label{sec:background}

Reinforcement learning (RL) is an optimization framework for tasks of a sequential nature~\citep{sutton1998reinforcement}. Typically, tasks are defined as a Markov decision process ($\mathcal{S}$, $\mathcal{A}$, $\mathcal{R}$, $p$, $d_0$, $\y$), with state space $\mathcal{S}$, action space $\mathcal{A}$, reward function $\mathcal{R}$, transition dynamics $p$, initial state distribution $d_0$, and discount factor $\y \in [0,1)$. Actions are selected according to a policy $\pi$~\citep{puterman1994markov}.

The performance of a policy is measured by its discounted return~$\E_\pi [ \sum_t^\infty \y^t r(s_t,a_t) ]$. Policy evaluation is the task of approximating the value function
~$Q^\pi(s,a) = \E_\pi [ \sum_t^\infty \y^t r(s_t,a_t) | s_0 = s, a_0 = a]$ of a target policy, given samples from an arbitrary dataset. %
A fundamental relationship regarding value functions is the Bellman equation~\citep{bellman, sutton1998reinforcement}: 
\begin{equation}
    Q^\pi(s,a) = \E_{r, s' \sim p,a' \sim \pi} \lb r + \y Q^\pi(s',a') \rb,
\end{equation}
which relates the value of the current state-action pair to an expectation over the next state-action pair. Given an approximate value function $Q$ (distinguished from the true value function $Q^\pi$ by dropping the $\pi$ superscript) of a target policy $\pi$, we denote the Bellman error $\e(s,a)$: %
\begin{equation}
    \label{eq:bellman-error}
    \e_Q(s,a) := Q(s,a) - \E_{r,s' \sim p,a' \sim \pi} \lb r + \y Q(s',a') \rb.
\end{equation}
In policy evaluation, the main objective of interest is a loss (such as the MSE or L1) over the value error~$\Delta_Q(s,a)$, the distance of an approximate value function $Q$ to the true value function~$Q^\pi$ of the target policy $\pi$: %
\begin{equation}
    \label{eq:value-error}
    \Delta_Q(s,a) := Q(s,a) - Q^\pi(s,a).
\end{equation}
Value error is often unmeasurable, as the true value function~$Q^\pi$ is unobtainable without the underlying MDP. While both the Bellman error and the value error are defined with respect to $Q$, for simplicity we drop the subscript when the error terms are not in reference to a specific value function. 

A standard result is the Bellman equation is uniquely solved by the true value function~\citep{bertsekas1995dynamic}. 
This can be re-framed in terms of Bellman and value error.  
\begin{proposition}
The Bellman error $\e(s,a)=0$ for all state-action pairs $(s,a) \in \mathcal{S} \times \mathcal{A}$, if and only if the value error $\Delta(s,a)=0$ for all state-action pairs $(s,a) \in \mathcal{S} \times \mathcal{A}$.
\end{proposition}
The Bellman equation can be used as an operator $\mathcal{T}Q(s,a) = \E_{r,s' \sim p,a' \sim \pi} \lb r + \y Q(s',a') \rb$, which when applied repeatedly to any function over all state-action pairs will converge to the true value function~\citep{puterman1994markov}. %
\begin{proposition} \label{prop:bellman_operator}
Let $Q_1, Q_2$ be value functions, where $Q_2 = \mathcal{T} Q_1$. $\max_{(s,a)} |\Delta_{Q_2}(s,a)| \leq \y \cdot \max_{(s,a)} |\Delta_{Q_1}(s,a)|$. 
\end{proposition}
In instances where we cannot compute the Bellman error exactly, 
we can instead use the temporal difference (TD) error $\delta(i)$, a sample-based approximation to the Bellman error which is computed over a transition~$i:=(s,a,r,s')$, $\delta(i) := Q(s,a) - (r + \y Q(s',a'))$, where $a' \sim \pi$ \citep{sutton1988tdlearning}. Note that the expected TD error is the Bellman error $\e(s,a) = \E_{r,s',a'}[\delta(i)]$, where the two errors are identical if the environment and policy are deterministic. 

A related concept is the projected Bellman error~\citep{sutton1998reinforcement, patterson2022generalized}. Since a given function approximator (particularly linear functions) may be unable to represent value functions with low Bellman error, the projected Bellman error is a projection into the space that is representable by the function approximator. In our work, we consider expressive deep neural networks which can fully represent low Bellman error solutions (as shown by our experiments in \autoref{sec:experiments}), allowing us to consider the true Bellman error, and avoid the need for projections.  

We focus largely on two algorithms based on the Bellman equation, which update an approximate value function $Q$, using samples from a finite dataset $\mathcal{D}$. 
Bellman residual minimization~(BRM)~\citep{baird1995residual} directly minimizes the Bellman error over samples in the dataset $\mathcal{D}$:
\begin{equation}
\label{eq:brm}
\Loss_\text{BRM}(Q) := \frac{1}{|\mathcal{D}|} \sum_{\mathcal{D}} ( Q(s,a) - (r + \y Q(s',a')) )^2.
\end{equation}
Fitted Q-Evaluation~(FQE)~\citep{ernst2005tree, le2019batch} is an iterative method with a similar update:
\begin{equation}
    \label{eq:fqe}
\Loss_\text{FQE}(Q) := \frac{1}{|\mathcal{D}|} \sum_{\mathcal{D}} ( Q(s,a) - (r + \y \bar Q(s',a')) )^2.
\end{equation}
The key distinction between the two algorithms is that BRM updates both $Q(s,a)$ and $Q(s',a')$, while FQE only updates $Q(s,a)$. This is because FQE uses $\bar Q(s',a')$, a fixed target value function updated by $\bar Q \leftarrow Q$ after a set number of time steps (possibly including every time step). 

\section{Bellman Error as a Proxy for Value Error} \label{sec:BE}

In this section, we analyze the relationship between the Bellman error and value error. Recall the key idea behind the Bellman equation is that it is uniquely satisfied by the true value function over all state-action pairs. %
In policy evaluation, the Bellman error of an approximate value function, computed over a set of state-action pairs, is used as a measurable proxy objective to the value error, the typically unmeasurable, true objective. 

While the Bellman error of the true value function is zero for all state-action pairs, in this section we show that the magnitude of the Bellman error of an approximate value function is only loosely related to the value error, even when considering the entire MDP.  
We show this breakdown by highlighting two fundamental problems with using the Bellman error as a surrogate for value error. 

\subsection{(Infinite Data Analysis) Problem 1: The Magnitude of the Bellman Error Hides Bias}\label{sec:problem1} 

The first problem is that the magnitude of the Bellman error is not guaranteed to correspond with the magnitude of the value error. This is because the Bellman error of a state-action pair is the difference between two values, which means there can exist cancellations in error terms, resulting in deceptively low Bellman error. 

To see this cancellation, consider the following:  %
\begin{observation}\label{eqn:BE_as_VE} {\normalfont \textbf{(Bellman error as a function of value error).}}
For any state-action pair $(s,a) \in \mathcal{S} \times \mathcal{A}$, the Bellman error~$\e(s,a)$ can be defined as a function of the value error~$\Delta$ 
\begingroup\abovedisplayskip=4pt \belowdisplayskip=3pt
\begin{equation}
\e(s,a) = \Delta(s,a) - \y \E_{s', \pi}[\Delta(s',a')].
\end{equation}
\endgroup
\end{observation}
Which follows directly from the Bellman equation: 
\begingroup\abovedisplayskip=4pt \belowdisplayskip=4pt
\begin{align}
    \e&(s,a) = Q(s,a) - \mathcal{T}Q(s,a) \\
    &= Q^\pi(s,a) + \Delta(s,a) - \mathcal{T}(Q^\pi(s,a) + \Delta(s,a)) \\
    &= \Delta(s,a) - \y \E_{s', \pi}[ \Delta(s',a') ].
\end{align}
\endgroup
While this does mean that smaller value error terms (in magnitude) does result in smaller Bellman error, the \textit{sign} of the value error terms can have a much larger impact. Keeping this relationship in mind, we can construct examples where the correspondence between the magnitude of the Bellman error and value error breaks down. 

\textbf{Example 1. (Same value error, different Bellman error).}  Let $Q^\pi$ be the true value function for some MDP and policy~$\pi$. We define the following approximate value functions: 
\begingroup\abovedisplayskip=6pt \belowdisplayskip=6pt
\begin{align}
    &Q_1 = Q^\pi + 1  &\text{(error is correlated),}\\
    &Q_2 = Q^\pi \pm 1  &\text{(error is uncorrelated),}
\end{align}
\endgroup
where $\pm 1$ is a random variable with equal probability of being positive or negative. In both cases, the absolute value error will be $1$ for any state-action pair. However, following \autoref{eqn:BE_as_VE}, for all state-action pairs, the Bellman error of $Q_1$ will be $1 - \y$, while the expected absolute Bellman error of $Q_2$ will be much larger at $|\pm 1 - \y \E[\pm 1]| = 1$. 

\textbf{Example 2. (Same Bellman error, different value error).} Suppose instead that we define $Q_1 = Q^\pi + \frac{1}{1 - \y}$. Then by \autoref{eqn:BE_as_VE}, the expected absolute Bellman error of $Q_1$ and $Q_2$ will both be $1$ for any state-action pair, but the value error of $Q_1$ will be $\frac{1}{1 - \y}$. For a common $\y=0.99$ and any state-action pair this would make the value error of $Q_1$ be $100$, compared to the value error of $Q_2$ at $1$. 

In fact, we can also define $Q_1$ and $Q_2$ such that the magnitude of the Bellman error and value error appear inversely correlated. 
\begin{proposition} {\normalfont \textbf{(An inverse relationship between Bellman error and value error).}}
For any MDP, discount factor $\y \in (0,1)$, and $C>0$, we can define a value function $Q_1$ and a stochastic value function $Q_2$ such that for any state-action pair~$(s,a)\in \mathcal{S} \times \mathcal{A}$
\begin{enumerate}[nosep]
    \item $|\Delta_{Q_1}(s,a)| - |\Delta_{Q_2}(s,a)| > C$,
    \item $\E_{Q_2}[|\e_{Q_2}(s,a)|] - |\e_{Q_1}(s,a)| > C$.
\end{enumerate}
\vspace{-2pt}
\end{proposition}
This means that when comparing value functions, lower absolute Bellman error over \textit{all} state-action pairs does not guarantee lower value error for \textit{any} state-action pair. %

However, there does exist a relationship between the value error of a given state-action pair, and the Bellman error of all relevant state-action pairs of an MDP. 
Let $d^\pi(s',a'|s,a)$ be the conditional discounted state-action occupancy $(1 - \y) \sum_{t=0}^\infty \y^t p^\pi((s,a) \rightarrow s', t) \pi(a'|s')$, 
where $p^\pi((s,a) \rightarrow s', t)$ is the probability of leaving the state-action pair $(s,a)$ and visiting the state $s$ after $t$ time steps. 
\begin{theorem} \label{thm:bellman_value}
{\normalfont \textbf{(Value error as a function of Bellman error).}}
For any state-action pair~$(s,a)\in \mathcal{S} \times \mathcal{A}$, the value error~$\Delta(s,a)$ can be defined as a function of the Bellman error~$\e$ 
\begingroup\abovedisplayskip=4pt \belowdisplayskip=4pt
\begin{equation}
    \Delta(s,a) = \frac{1}{1 - \y} \E_{(s',a') \sim d^\pi(\cdot|s,a)} [\e(s',a')].
\end{equation}
\endgroup
\end{theorem}
\autoref{thm:bellman_value} provides an upper bound on the absolute value error by considering the expected absolute Bellman error of future state-action pairs (according to $d^\pi$)
\begingroup\abovedisplayskip=6pt \belowdisplayskip=6pt
\begin{equation} \label{eqn:upperbound}
|\Delta(s,a)| \leq \frac{1}{1 - \y} \E_{(s',a') \sim d^\pi(\cdot|s,a)} [|\e(s',a')|].
\end{equation}
\endgroup
Since the right-hand side of this bound corresponds to the on-policy Bellman residual minimization objective with a L1 loss, this shows how the magnitude of on-policy Bellman error can be used as an objective for policy evaluation. 
However, we remark that this also highlights how the absolute Bellman error can correspond to a wide range of possible value error terms by also considering the lower bound. %
Let $d^\pi$ be the expected discounted state-action occupancy $\E_{s_0}[d^\pi(s,a|s_0)]$ over the initial state distribution. 
\begin{proposition}\label{prop:bounds}
{\normalfont \textbf{(Bounds on value error).}}
Let $C_\text{max}=\max_{s,a} |\e(s,a)|$ and $C_\text{avg}=\E_{(s,a) \sim d^\pi} [|\e(s,a)|]$. 
\begin{enumerate}[nosep]
    \item $\frac{C_\text{max}}{1+\y} \leq \max_{s,a} |\Delta(s,a)| \leq \frac{C_\text{max}}{1-\y}$,
    \item $\frac{C_\text{avg}}{1+\y} \leq \E_{(s,a) \sim d^\pi} [|\Delta(s,a)|] \leq \frac{C_\text{avg}}{1-\y}$,
\end{enumerate}
Furthermore, for any MDP and policy, there exists a value function such that the upper bound is an equality, and there exists a MDP, policy, and value function such that the lower bound is an equality. 
\end{proposition}
Similarly to Example 2, \autoref{prop:bounds} says that if the average absolute Bellman error is $1$, and the discount factor $\y=0.99$, then the average value error will fall somewhere between $\approx 0.5$ and $100$. Consequently, this wide range of value error suggests that it is difficult to make meaningful comparisons between value functions using the magnitude of the Bellman error alone.

\begin{figure}
\centering
\small
\begin{tikzpicture}[
indata/.style={draw,circle, minimum size=30pt, very thick}, %
outdata/.style={draw,circle, minimum size=30pt, very thick}
]
\node[indata] (s0) at (0,0) {$s_0$};
\node[outdata] (s1) at (2.5,0) {$s_1$};

\draw[-{Latex[length=5pt]}, thick] (s0) -- node[above]{$a$} (s1);
\draw[-{Latex[length=5pt]}, thick] (s1.-25) arc (222:270+222:10pt) node[midway,right]{$a$};
\end{tikzpicture}
\caption{A basic MDP. If $(s_0,a)$ is contained in the dataset but $(s_1,a)$ is not, by selecting the values $Q(s_0,a)$ and $Q(s_1,a)$, we can construct examples where the Bellman error over the dataset is $0$, but the value error is arbitrarily large, and vice-versa.  %
} \label{fig:basic_mdp}
\vspace{-8pt}
\end{figure}
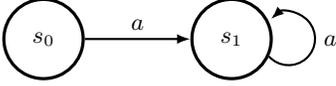

\vspace{-2pt}
\subsection{(Finite Data Analysis) Problem 2: Missing Transitions Breaks the Bellman Equation}\label{sec:problem2} 

A direct consequence of \autoref{thm:bellman_value} is the uniqueness property of the Bellman equation. That is, if the Bellman error is~$0$ for all relevant state-action pairs which may be visited by the target policy, then the value error must also be~$0$ for these state-action pairs.  
However, if we examine a finite dataset instead of the entire MDP, this relationship also %
highlights that if any relevant transitions are missing, the Bellman equation, evaluated over the dataset, will no longer have a unique solution. 
This means that when considering a finite dataset, there can exist infinitely many suboptimal solutions which satisfy the Bellman equation exactly. 

\begin{corollary} \label{corollary:not_unique}
{\normalfont \textbf{(The Bellman equation is not unique over incomplete datasets).}}
If there exists a state-action pair $(s',a')$ not contained in the dataset $\mathcal{D}$, where the state-action occupancy $d^\pi(s',a'|s,a) > 0$ for some $(s,a) \in \mathcal{D}$, then there exists a value function and $C>0$ such that 
\begin{enumerate}[nosep]
    \item For all $(\hat s, \hat a) \in \mathcal{D}$, the Bellman error $\e(\hat s,\hat a) = 0$.
    \item There exists $(s,a) \in \mathcal{D}$, such that the value error $\Delta(s,a) = C$.
\end{enumerate}
\vspace{-2pt}
\end{corollary}
We can find examples of this phenomenon in very simple MDPs. Consider the two-state MDP defined in \autoref{fig:basic_mdp}, with reward $r=0$ for all state-action pairs. If the dataset contains the sole transition~$(s_0,a,r,s_1)$, then we can construct examples where for all transitions in the dataset (the sole transition in this instance) the Bellman error is $0$ but the value error is arbitrarily large, and conversely, where the Bellman error is arbitrarily large but the value error is~$0$. 

\textbf{\setword{Example 1}{example1}. ($0$ Bellman error, $C$ value error).} Since the reward is $0$ everywhere, the true value of any state-action pair must be $0$. However, by choosing the value of the missing transition $Q(s_1, a)$ to cancel with the value of the transition in the dataset $Q(s_0, a)$, the Bellman error of the sole state-action pair in the dataset is $0$. 
\begin{equation}
    \begin{aligned} 
    Q(s_0,a)&=C,\\ 
    Q(s_1,a)&=\textstyle \frac{1}{\y} C. 
    \end{aligned} 
    \implies
    \begin{aligned}
    \e(s_0,a)&= C - \y \textstyle \frac{1}{\y} C = 0,\\
    \Delta(s_0,a)&= Q(s_0,a) = C.
    \end{aligned}
\end{equation}
\textbf{\setword{Example 2}{example2}. ($C$ Bellman error, $0$ value error).} Similarly, we can set the value of the state-action pair in the dataset $Q(s_0, a)$ to be the correct value of $0$, and instead choose the value of the missing state-action pair $Q(s_1, a)$ to make the Bellman error of $(s_0,a)$ equal $C$.
\begin{equation}
    \begin{aligned} 
    Q(s_0,a)&=0,\\ 
    Q(s_1,a)&= - \textstyle \frac{1}{\y} C.
    \end{aligned} 
    \implies \hspace{-4pt}
    \begin{aligned}
    \e(s_0,a)&= 0 + \y \textstyle \frac{1}{\y} C = C,\\
    \Delta(s_0,a)&= Q(s_0,a) = 0.
    \end{aligned}
\end{equation}
Note that these examples do not involve modifying the environment in an extreme adversarial manner, and instead occur due to the value estimate of the missing state-action pair. As a result, methods which directly minimize the Bellman error may influence target values to achieve low Bellman error over the dataset without necessarily improving the value function. Even when the target value is not directly modified, scenarios such as \ref{example1}, where the current and missing values are similar, can cause deceptively low Bellman error, without an accurate value function.

\begin{figure*}[t]
    \centering
    \includegraphics[width=\textwidth]{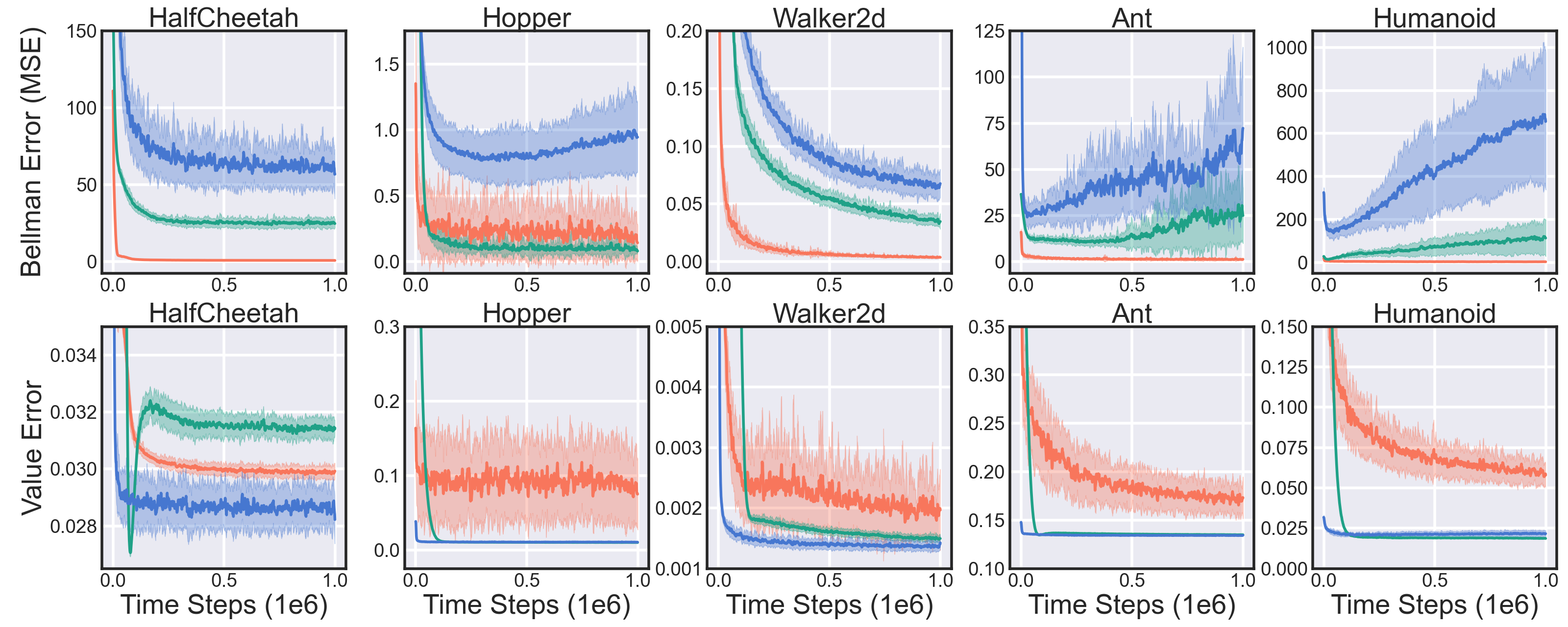}
    \small \cblock{my_warm_2} BRM \qquad \cblock{my_cold_3} FQE \qquad \cblock{my_blue} MC
    \caption[]{\textbf{Can the on-policy Bellman error be used as a proxy for value error?} We compare the mean squared Bellman error (top row) with the normalized absolute value error (bottom row) of learning curves of value functions trained with a dataset of 1m \textit{on-policy} transitions. 
    While clearly the Bellman error is lowest for\cblock{my_warm_2}~BRM (which directly minimizes Bellman error) followed by\cblock{my_cold_3}~FQE~(which indirectly minimizes Bellman error) followed by\cblock{my_blue}~MC~(which minimizes the Monte-Carlo estimate of value error), this ordering is not reflected in value error. This shows overfitting of the Bellman error objective is possible even with on-policy data and that we cannot determine value prediction accuracy by examining the magnitude of Bellman error alone.}
    \label{fig:both}
\end{figure*}

\begin{figure}[t]
    \centering
    \includegraphics[width=0.49\linewidth]{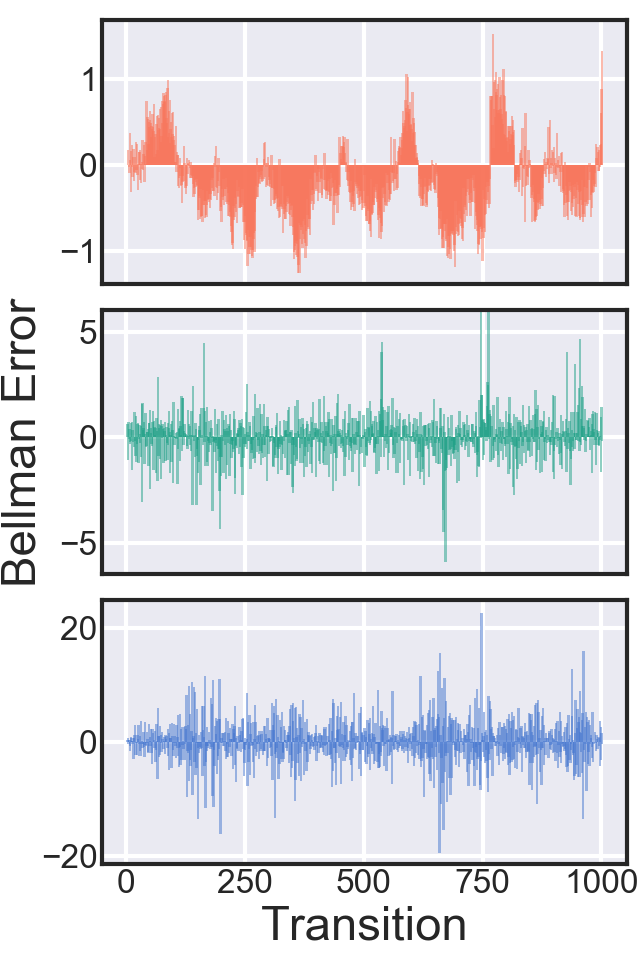}
    \includegraphics[width=0.49\linewidth]{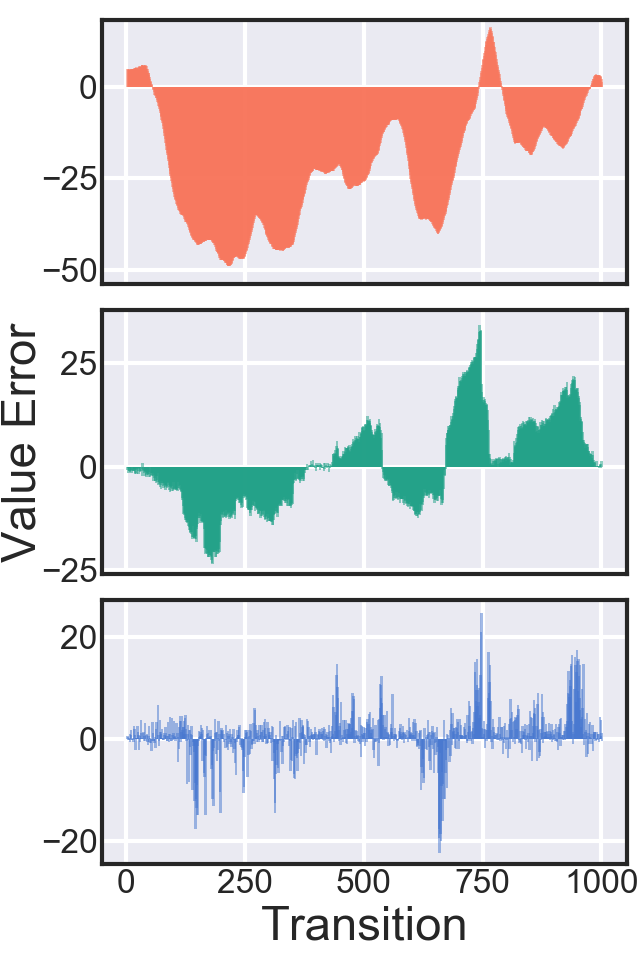}
    \small \cblock{my_warm_2} BRM \qquad \cblock{my_cold_3} FQE \qquad \cblock{my_blue} MC
    \setlength{\tabcolsep}{4.5pt}
    \vspace{5pt}
    
    \begin{tabular}{lrrr}
    \toprule
        & Mean $|\text{BE}|$   & Mean $|\text{VE}|$ & Ratio $(\sum |\text{VE}|/ \sum |\text{BE}|)$ \\ \midrule
    BRM & 0.41 & 23.39 & 56.93\\
    FQE & 0.76 & 8.89 & 11.72\\
    MC  & 2.76 & 2.75 & 1.00\\ \bottomrule
    \end{tabular}
    \caption{\textbf{Does bias explain the non-correspondence between the Bellman error and value error?} We train value functions over a single on-policy trajectory in the HalfCheetah environment (1000 transitions, with termination). As a result, every relevant transition is contained in the dataset. We display the Bellman error (left) and unnormalized value error (right) for every transition in the trajectory. In the table, we record the average absolute Bellman error (BE), value error (VE), and ratio between these terms. By \autoref{prop:bounds}, with a discount $\y=0.99$, the ratio between mean absolute Bellman and value error is bounded by $[0.503, 100]$. Our results show that both the upper and lower bound can be approached in practice. 
    Visually, we can see that the magnitude of the Bellman error is influenced more by the clusters of similar value error (size and direction) for\cblock{my_warm_2}~BRM and\cblock{my_cold_3}~FQE than the magnitude of the value error. This experiment is repeated with different domains in Appendix \ref{appendix:one_traj}. 
    } \label{fig:one_traj}
    \vspace{-8pt}
\end{figure}

\section{Experiments} \label{sec:experiments}

Everything we have discussed thus far has suggested that the Bellman error may not be a representative surrogate for value error. We now examine if these problems occur in practice. Recall our main observations are (1) the magnitude of the Bellman error is smaller for biased value functions due to cancellations caused from both sides of the Bellman equation (\autoref{sec:problem1}) and (2) the relationship between Bellman error and value error is broken if the dataset is missing relevant transitions (\autoref{sec:problem2}). We find that both of these concerns exist clearly in standard benchmark environments.

Our experiments also demonstrate the effectiveness of Fitted Q-Evaluation (FQE)~\citep{ernst2005tree, le2019batch} as a policy evaluation method, irrespective of its tendency to find value functions with high Bellman error. While FQE is also based on the Bellman equation, we argue that its objective is not to directly minimize the Bellman error, but rather to imitate the Bellman operator, meaning that low error to an accurate target can be a more important factor than the Bellman error~(\autoref{sec:FQE}). 

\subsection{Experimental Design}

We first outline the experimental choices used in our empirical evaluation. Our experiments consider the setting of policy evaluation. Comprehensive details can be found in \autoref{appendix:details}. 
The task can be broken down as follows:
\begin{enumerate}[nosep]
    \item The target policy is generated from an off-the-shelf RL algorithm. 
    \item The behavior policy collects a training and test dataset.
    \item The value function is trained using the dataset, target policy, and policy evaluation algorithm.
    \item Metrics of the value function are computed over the test dataset.
\end{enumerate}

\textbf{Policies.} The target policy is an expert deterministic policy from a fully trained TD3 agent~\citep{fujimoto2018addressing}. The behavior policy is a noisy version of the target policy. Each dataset is assigned a noise value corresponding to both the probability of selecting a uniformly random action, as well as the standard deviation of Gaussian noise added to the actions (noting that actions are normalized to be in the range $[-1,1]$). We use uniformly random actions to ensure that not all actions in the dataset are centered around the target policy, and Gaussian noise to ensure that every action is distinct from actions selected by the target policy. The on-policy behavior is generated without noise. 

\textbf{Algorithms.} We use Bellman residual minimization (BRM) \citep{baird1995residual}, Fitted Q-Evaluation~(FQE) \citep{ernst2005tree, le2019batch}, and Monte Carlo estimates (MC) \citep{sutton1998reinforcement}. We use these algorithms due to their prevalence in the literature, and to highlight differences between methods which minimize the Bellman error, directly (BRM) or indirectly (FQE), and methods which minimize value error~(MC).

\textbf{Metrics.} We compare the mean squared Bellman error, due to its common use~\citep{baird1995residual, sutton1998reinforcement}, with the absolute value error. For better interpretability across tasks, we normalize the value error by dividing by a constant term equal to the average true value function sampled on-policy. The true value function is estimated with on-policy rollouts from the target policy. In Appendix \ref{appendix:absabs} we report results with variations of these metrics.

\textbf{Environments.} Our experiments consider standard benchmark continuous-action tasks through the MuJoCo simulator~\citep{mujoco, OpenAIGym}, as it is deterministic and high-dimensional. Determinism in the dynamics is desirable as it, alongside a deterministic policy, makes the TD error identical to the Bellman error. This means we can compute the Bellman error exactly and ignore the double sampling issue~\citep{baird1995residual}.

\subsection{The On-Policy Bellman Error is a Poor Metric} \label{sec:problem1_experiments}

In \autoref{sec:problem1}, we showed that it is possible for a biased value function to have lower Bellman error over all possible state-action pairs, while having higher value error, than an unbiased value function. %
In this section we show that this problem exists in practical settings. This means that the magnitude of Bellman error is a poor metric for comparing value functions, even when working with on-policy data.

We begin by comparing the Bellman error and the value error of value functions trained, and evaluated, with on-policy data. 
The results are displayed in \autoref{fig:both}. %

\textbf{Methods which minimize the Bellman error have lower Bellman error.} BRM directly minimizes the Bellman error, FQE uses an iterative approach based on the Bellman equation, and MC does not use the Bellman equation in its objective. Therefore it is unsurprising that in \autoref{fig:both}, the value functions trained by BRM have the lowest average Bellman error, followed by FQE, and then MC.

\textbf{Lower Bellman error does not mean lower value error.} Although the value functions trained by FQE and MC have much higher average absolute Bellman error over the on-policy dataset~(especially in Ant and Humanoid) than BRM, 
the value functions trained by BRM typically have the highest average value error.  
The learning curves in \autoref{fig:both} demonstrate that it is possible to find value functions where the relative ordering in Bellman error is not respected by the relative ordering in value error, even when working with large (1M) on-policy datasets. 

\textbf{Methods which minimize the Bellman error have correlated error} (and thereby have a higher ratio of value error to Bellman error).  
To better understand the importance of bias in minimizing the Bellman error, we train value functions over a single trajectory. This allows us to consider a dataset without missing transitions and examine the error terms of every transition in the trajectory. We display the results in \autoref{fig:one_traj}. As the Bellman error is determined by the difference between the value error of the current and succeeding state (\autoref{eqn:BE_as_VE}), \autoref{fig:one_traj} shows the magnitude of the Bellman error is influenced more by correlated bias than the magnitude the value error. 

\newcommand{\gc}{{\color{sb_green}\checkmark}}

\begin{table*}[t]
\setlength{\tabcolsep}{6pt}
\small
\centering
\begin{tabular}{ccrr@{\hspace{1pt}}lrr@{\hspace{1pt}}lrr@{\hspace{1pt}}lrr@{\hspace{1pt}}lrr@{\hspace{1pt}}l}
\toprule
Algorithm & Dataset & \multicolumn{3}{r}{HalfCheetah\hspace{7pt}~} & \multicolumn{3}{r}{Hopper\hspace{15pt}~} & \multicolumn{3}{r}{Walker2d\hspace{16pt}~} & \multicolumn{3}{r}{Ant\hspace{20pt}~} & \multicolumn{3}{r}{Humanoid\hspace{10pt}~} \\ \midrule
&         & BE               & VE &          & BE            & VE &        & BE            & VE &          & BE          & VE &       & BE             & VE &         \\
\midrule
\multirow{6}{*}{BRM}          & On-Policy       & 0.96             & 0.03 & \gc       & 0.17          & 0.08 & \gc     & {\tiny \raisebox{0.2em}{\textless}} 0.01        & {\tiny \raisebox{0.2em}{\textless}} 0.01 & \gc     & 2.54        & 0.16 & \gc    & 1.11           & 0.06 & \gc      \\
                              & 0.1     & 2.02             & 0.36 &        & 3.50          & 0.47 &      & 2.10          & 0.37 &        & 5.36        & 0.81 &     & 0.53           & 0.69 &       \\
                              & 0.2     & 0.70             & 0.47 &       & 1.86          & 0.61 &      & 1.07          & 0.45 &       & 1.27        & 0.68 &     & 0.28           & 0.80 &       \\
                              & 0.3     & 0.34             & 0.49 &        & 1.11          & 0.62 &      & 0.55          & 0.43 &        & 0.63        & 0.62 &     & 0.16           & 0.70 &       \\
                              & 0.4     & 0.18             & 0.48 &        & 1.08          & 0.60 &      & 0.33          & 0.41 &        & 0.38        & 0.62 &     & 0.13           & 0.63 &       \\
                              & 0.5     & 0.09             & 0.50 &        & 0.23          & 0.63 &      & 0.26          & 0.43 &        & 0.35        & 0.56 &     & 0.07           & 0.63 &       \\
\midrule
\multirow{6}{*}{FQE}          & On-Policy       & 38.96            & 0.03 & \gc       & 0.09          & 0.01 & \gc     & 0.05          & {\tiny \raisebox{0.2em}{\textless}} 0.01  &    \gc & 37.95       & 0.11 & \gc    & 45.58          & 0.02 & \gc      \\
                              & 0.1     & 1375.71          & 0.13 & \gc       & 17.56         & 0.03 & \gc     & 88.75         & 0.04 & \gc       & 838.39      & 0.11 & \gc    & 137.83         & 0.03 & \gc      \\
                              & 0.2     & 2441.75          & 0.12 & \gc        & 57.24         & 0.05 & \gc     & 147.77        & 0.07 & \gc        & 984.95      & 0.14 & \gc    & 342.16         & 0.06 & \gc      \\
                              & 0.3     & 1219.08          & 0.18 &        & 95.72         & 0.07 & \gc     & 113.13        & 0.11 &        & 708.50      & 0.19 & \gc    & 633.68         & 0.09 & \gc      \\
                              & 0.4     & 244.47           & 0.21 &        & 214.92        & 0.40 &      & 139.46        & 0.12 &        & 522.05      & 0.26 &     & 575.56         & 0.09 & \gc      \\
                              & 0.5     & 144.04           & 0.25 &        & 169.78        & 0.22 &      & 35.38         & 0.21 &        & 343.37      & 0.30 &     & 483.69         & 0.10 & \gc      \\ 
\bottomrule
\end{tabular}
\caption{\textbf{Is the Bellman error an effective off-policy objective?} We display the final mean squared Bellman error (BE) and the final normalized absolute value error (VE) of functions trained with datasets gathered by increasingly noisy versions of the target policy. The value of the dataset refers to the noise level. \gc~is used to mark if performance is within $0.1$ normalized value error of the on-policy value error. 
FQE consistently outperforms BRM while having significantly higher Bellman error, particularly when off-policy. Additionally, while the value error of BRM increases substantially from on-policy to off-policy datasets, the Bellman error remains low for all datasets, showing that it is possible to train a function with low Bellman error, regardless of its accuracy with respect to the true value function.} \label{table:OPE_obj}
\end{table*}

\textbf{The Bellman error is a valid on-policy objective.} 
In Appendix~\ref{appendix:correlation} we display the correlation coefficient between the average absolute Bellman error and value error, when isolating functions by the algorithm which trained them. We find that when the error terms are evaluated on-policy, there is a strong correlation between the error terms for a fixed algorithm. 
This corroborates with \autoref{eqn:upperbound} which shows that the expected absolute Bellman error of on-policy samples upper bounds value error, and suggests that when we isolate by algorithm, value functions are biased similarly. This means that even if the Bellman error is a poor metric for comparison, the Bellman error can be a usable objective for on-policy evaluation and explains the decent value prediction accuracy of BRM in \autoref{fig:both}. 

\textbf{Variations of the Bellman error as a metric.}  
In \autoref{sec:problem1}, we showed the magnitude of the Bellman error hides bias and is therefore a poor metric for comparing value functions. Conceivably, one could incorporate a measure of bias into the Bellman error. For example, following \autoref{thm:bellman_value}, the exact value error can be computed from on-policy data.  %
However, when working with incomplete datasets we can always find functions which have zero Bellman error and arbitrarily high value error for any state-action pair in the dataset~(\autoref{corollary:not_unique}). Consequently, there cannot exist a metric without failure cases for finite off-policy datasets.

\subsection{The Off-Policy Bellman Error is a Poor Objective} \label{sec:problem2_experiments}

In the previous section, we showed that the Bellman error cannot be used  
to rank arbitrary value functions according to their accuracy. Regardless, the on-policy accuracy of BRM is competitive (albeit weaker) with FQE and MC. 
This suggests the possibility that the Bellman error could be used as an objective for off-policy evaluation as well, irrespective of its shortcomings as a metric. In this section we show that this is not the case: the Bellman error makes for a poor off-policy objective.

Recall in \autoref{sec:problem2} we showed that if there are missing transitions in the dataset, the Bellman equation can be satisfied exactly by infinitely many suboptimal solutions. This is largely a problem for finite datasets generated off-policy, as they are guaranteed to be missing relevant transitions. 
Consequently, the Bellman error, evaluated over an off-policy dataset, should not be a meaningful objective for policy evaluation. In \autoref{table:OPE_obj} we confirm this observation, by repeating the policy evaluation experiment from the previous section with various off-policy datasets. 

\textbf{The Bellman error is an ineffective off-policy objective.} \autoref{table:OPE_obj} shows the performance of BRM suffers a clear drop when moving to the off-policy datasets. Furthermore, despite this performance drop, the average Bellman error of value functions trained by BRM remains low for all datasets, which shows that the empirical Bellman error, evaluated over an off-policy dataset, can be optimized independently of value error. Moreover, we remark that our experiments are in deterministic domains, and as such, this problem is independent from the double sampling problem with BRM~\citep{baird1995residual}. These results support our theoretical analysis (\autoref{corollary:not_unique}) which showed that with missing transitions, the Bellman error is no longer guaranteed to relate to value error. 

We also include results of the FQE algorithm on these datasets. While FQE also suffers in performance with additional noise added to the behavior policy, this performance drop is much less drastic. On every off-policy dataset, FQE outperforms BRM, and in some instances, even outperforms BRM trained with the on-policy dataset.  %
The success of FQE shows that the failure of BRM is due to its objective (of Bellman error), rather than the difficulty of the task.

\vspace{-4pt}
\subsection{Additional Insights about FQE} \label{sec:FQE}

The success of FQE for policy evaluation is supported by many examples in the literature~\citep{voloshin2021empirical, fu2021benchmarks, fujimoto2021srdice}, as well as control~\citep{DQN, DDPG, hessel2018rainbow}. The results in the previous sections may raise new questions surrounding the performance, and behavior, of FQE. 

\textbf{Does FQE provide a better metric?} In Appendix~\ref{appendix:correlation_FQE} we examine the relationship of the FQE loss function to value error. This loss bears no closer relationship to value error than the Bellman error does. This is not contradictory to the success of FQE because the objective is iterative and evolves during training. We can interpret the objective of FQE as an application of the Bellman operator, 
where the Bellman equation is used to construct a regression target, rathar than a minimization of the Bellman error. This means that FQE does not provide a better metric than the Bellman error, only a better optimization process.

\textbf{How does FQE handle missing data?} The success of FQE is best understood as the repeated application of the Bellman operator. Recall the Bellman operator, applied globally to every state-action pair, ensures a reduction in the max value error (\autoref{prop:bellman_operator}). We can instead consider the Bellman operator applied over a finite dataset~$\mathcal{D}$. 
\begin{proposition}\label{prop:fqe_contraction}
{\normalfont \textbf{(FQE improvement condition).}} Let $Q_1, Q_2$ be value functions, where $Q_2 = \mathcal{T} Q_1$. 
If $\y \cdot \max_{(s,a) \in \mathcal{D}} |\E_{(s',a') \sim p(\cdot|s,a), \pi}[\Delta_{Q_1}(s',a')]| < \max_{(s,a) \in \mathcal{D}} |\Delta_{Q_1}(s,a)|$ then $\max_{(s,a) \in \mathcal{D}} |\Delta_{Q_2}(s,a)| < \max_{(s,a) \in \mathcal{D}} |\Delta_{Q_1}(s,a)|$.
\end{proposition}
Informally, \autoref{prop:fqe_contraction} says that if the target used by FQE is sufficiently accurate relative to the accuracy of the value function over state-action pairs in the dataset, then an iteration of FQE will improve the value estimate over the dataset. This means that FQE can overcome missing data through generalization and explains the strong performance of FQE in some of the off-policy settings in \autoref{table:OPE_obj}. BRM on the other hand, inhibits generalization in the target, as the target can be directly modified to reduce Bellman error, which can result in overfitting.  

\textbf{FQE relies on generalization during training.} While the importance of generalization to FQE is a simple observation, it has significant implications. Firstly, this means the Bellman equation requires generalization \textit{during} training. This is distinct from typical machine learning settings, where generalization is an exercise which occurs \textit{after} training. This is problematic because if it is difficult to ensure good generalization after training, it is only more difficult to ensure good generalization during training. This highlights the importance of feature learning~\citep{jaderberg2016reinforcement, yang2021representation}, as neural network features are unlikely to be relevant early in training. 

Another implication is hyperparameter sensitivity. It is a well known problem that RL algorithms are sensitive to small adjustments~\citep{henderson2017deep, engstrom2019implementation}. A necessity of generalization at training time causes the significance of correct hyperparameters to be amplified. 

\vspace{-2pt}
\section{Related Work}

The role of the Bellman error has been considered in depth in the literature, in the context of bounds on the performance of a greedy policy in relation to the norm of the Bellman error
~\citep{williams1993tight, singh1994upper, bertsekas1995dynamic, heger1996loss, munos2003error, munos2007performance, farahmand2010error}.  

In a similar vein to our work, \citet{kolter2011fixed} remarks that with off-policy sampling, the projected solution to linear TD can have arbitrarily large Bellman error, but focuses on the approximate fixed point, rather than any discrepancy between Bellman error and value error. %
\citet{geist2017bellman} evaluate the Bellman error as an objective for policy optimization. Although they examine a different setting, they arrive at a related conclusion, the signal from the Bellman error is only meaningful if the sampling distribution corresponds to the optimal policy.

The Bellman error has additional concerns that our paper does not discuss. The double sampling problem~\citep{baird1995residual} is that the gradient of the Bellman error is biased if estimated from a single transition in a stochastic MDP. The double sampling issue provides motivation for many recent BRM methods~\citep{feng2019kernel, zhu2020borrowing, zhang2020deep, bas2021logistic}.
We avoid this issue in our analysis by focusing on deterministic environments, but remark that BRM is likely to perform worse with stochasticity. 
\citet{sutton1998reinforcement} show that in scenarios where the %
representation of states is not uniquely defined, there exist examples where the Bellman error is not learnable, as the structure of the MDP can not be determined from data alone, and thus the true Bellman error cannot be computed.

The success of FQE is supported by many examples in the literature for off-policy evaluation tasks~\citep{voloshin2021empirical, fu2021benchmarks, fujimoto2021srdice}, as well as control applications with deep RL~\citep{DQN, DDPG, hessel2018rainbow}. Our results help explain the effectiveness of FQE, but also highlights challenge of the importance of generalization during training. 

Our observations connect strongly to offline RL \citep{lange2012batch, levine2020offline}, where offline policy evaluation is used in conjunction with policy learning. 
Previous work has observed that %
the value function of FQE methods can diverge when computed offline due to poor estimates in the target~\citep{fujimoto2018off, fujimoto2019benchmarking}. Similar to our work, empirical properties of deep value functions which induce instability or divergence have been studied~\citep{fu2019diagnosing, achiam2019towards} but have not considered the role of the objective itself.
Several recent papers examined the sample complexity of offline RL, noting that without access to on-policy data, the number of necessary transitions is exponential with respect to the horizon~\citep{wang2020statistical, zanette2021exponential, chen2021infinite, xiao2021sample}. 

In the context of offline model selection, several papers have observed that TD error is a weak baseline with poor correspondence to policy performance, remarking that it is a measure of value function accuracy rather than quality of the policy, but provide little analysis~\citep{irpan2019off, paine2020hyperparameter, tang2021model}. Our findings help explain these results by showing that Bellman (and TD) error are not an effective measure of value accuracy, and cannot rank models, even with on-policy data. These empirically minded observations caution against traditional results which suggest Bellman error as a metric for model selection~\citep{farahmand2011model}.

\section{Conclusion}

In this paper, we study the effectiveness of the Bellman error as a proxy for value error. %
We focus on two problems. Firstly, the magnitude of the Bellman error is lower for biased value functions, as there are cancellations for temporally correlated error terms. This means that the magnitude of the Bellman error is not viable metric for selecting functions with lower value error, even when considering all state-action pairs. Secondly, the Bellman equation is only uniquely solved by the true value function when computed over the entire MDP. When working with finite or incomplete datasets, we show there exists infinitely many suboptimal value functions which satisfy the Bellman equation. This means the Bellman error is not a viable objective over incomplete datasets, such as the off-policy setting. 

We demonstrate these problems theoretically, with toy problems, and empirically on standard benchmark environments. The lack of correlation between Bellman error and value error is highlighted by an empirical comparison between Bellman Residual Minimization~(BRM)~\citep{baird1995residual} and Fitted Q-Evaluation~(FQE)~\citep{ernst2005tree, le2019batch}, which shows that value functions trained with BRM consistently have much lower Bellman error, but much higher value error than value functions trained with FQE. 

While much of the modern literature surrounding Bellman error minimization emphasizes the double sampling problem~\citep{dai2018sbeed, feng2019kernel, saleh2019deterministic, bas2021logistic}, our analysis shows much more fundamental problems; Bellman error is not guaranteed to closely correspond with value error and solving the Bellman equation over a finite dataset does not guarantee an accurate value function. 
While such problems may be alluded to in prior work~\citep{kolter2011fixed, geist2017bellman, irpan2019off}, to the best of our knowledge we are the first to formalize theorems and propositions in both infinite and finite data regimes, list out illustrative examples, and evaluate empirically on practical settings. We hope our findings provide a better understanding of Bellman equation-based objectives to both practitioners and theorists alike.

\section*{Acknowledgements}
Scott Fujimoto is supported by a NSERC scholarship. We would like to thank Dale Schuurmans, Pablo Samuel Castro, Esther Ugolini, Edward Smith, and Wei-Di Chang for helpful feedback and discussions.

\bibliography{example_paper}

\begin{thebibliography}{62}
\providecommand{\natexlab}[1]{#1}
\providecommand{\url}[1]{\texttt{#1}}
\expandafter\ifx\csname urlstyle\endcsname\relax
  \providecommand{\doi}[1]{doi: #1}\else
  \providecommand{\doi}{doi: \begingroup \urlstyle{rm}\Url}\fi

\bibitem[Achiam et~al.(2019)Achiam, Knight, and Abbeel]{achiam2019towards}
Achiam, J., Knight, E., and Abbeel, P.
\newblock Towards characterizing divergence in deep q-learning.
\newblock \emph{arXiv preprint arXiv:1903.08894}, 2019.

\bibitem[Antos et~al.(2008)Antos, Szepesv{\'a}ri, and Munos]{antos2008learning}
Antos, A., Szepesv{\'a}ri, C., and Munos, R.
\newblock Learning near-optimal policies with bellman-residual minimization
  based fitted policy iteration and a single sample path.
\newblock \emph{Machine Learning}, 71\penalty0 (1):\penalty0 89--129, 2008.

\bibitem[Badia et~al.(2020)Badia, Piot, Kapturowski, Sprechmann, Vitvitskyi,
  Guo, and Blundell]{badia2020agent57}
Badia, A.~P., Piot, B., Kapturowski, S., Sprechmann, P., Vitvitskyi, A., Guo,
  Z.~D., and Blundell, C.
\newblock Agent57: Outperforming the atari human benchmark.
\newblock In \emph{International Conference on Machine Learning}, pp.\
  507--517. PMLR, 2020.

\bibitem[Baird(1995)]{baird1995residual}
Baird, L.
\newblock Residual algorithms: Reinforcement learning with function
  approximation.
\newblock In \emph{Machine Learning Proceedings 1995}, pp.\  30--37. Elsevier,
  1995.

\bibitem[Bas-Serrano et~al.(2021)Bas-Serrano, Curi, Krause, and
  Neu]{bas2021logistic}
Bas-Serrano, J., Curi, S., Krause, A., and Neu, G.
\newblock Logistic q-learning.
\newblock In \emph{International Conference on Artificial Intelligence and
  Statistics}, pp.\  3610--3618. PMLR, 2021.

\bibitem[Bellman(1957)]{bellman}
Bellman, R.
\newblock \emph{Dynamic Programming}.
\newblock Princeton University Press, 1957.

\bibitem[Bertsekas \& Tsitsiklis(1996)Bertsekas and
  Tsitsiklis]{bertsekas1995dynamic}
Bertsekas, D.~P. and Tsitsiklis, J.~N.
\newblock \emph{Neuro-Dynamic Programming}.
\newblock Athena scientific Belmont, MA, 1996.

\bibitem[Bradtke \& Barto(1996)Bradtke and Barto]{bradtke1996linear}
Bradtke, S.~J. and Barto, A.~G.
\newblock Linear least-squares algorithms for temporal difference learning.
\newblock \emph{Machine learning}, 22\penalty0 (1):\penalty0 33--57, 1996.

\bibitem[Brockman et~al.(2016)Brockman, Cheung, Pettersson, Schneider,
  Schulman, Tang, and Zaremba]{OpenAIGym}
Brockman, G., Cheung, V., Pettersson, L., Schneider, J., Schulman, J., Tang,
  J., and Zaremba, W.
\newblock Openai gym, 2016.

\bibitem[Chen et~al.(2021)Chen, Scherrer, and Bartlett]{chen2021infinite}
Chen, L., Scherrer, B., and Bartlett, P.~L.
\newblock Infinite-horizon offline reinforcement learning with linear function
  approximation: Curse of dimensionality and algorithm.
\newblock \emph{arXiv preprint arXiv:2103.09847}, 2021.

\bibitem[Dai et~al.(2018)Dai, Shaw, Li, Xiao, He, Liu, Chen, and
  Song]{dai2018sbeed}
Dai, B., Shaw, A., Li, L., Xiao, L., He, N., Liu, Z., Chen, J., and Song, L.
\newblock Sbeed: Convergent reinforcement learning with nonlinear function
  approximation.
\newblock In \emph{International Conference on Machine Learning}, pp.\
  1125--1134. PMLR, 2018.

\bibitem[Engstrom et~al.(2019)Engstrom, Ilyas, Santurkar, Tsipras, Janoos,
  Rudolph, and Madry]{engstrom2019implementation}
Engstrom, L., Ilyas, A., Santurkar, S., Tsipras, D., Janoos, F., Rudolph, L.,
  and Madry, A.
\newblock Implementation matters in deep rl: A case study on ppo and trpo.
\newblock In \emph{International Conference on Learning Representations}, 2019.

\bibitem[Ernst et~al.(2005)Ernst, Geurts, and Wehenkel]{ernst2005tree}
Ernst, D., Geurts, P., and Wehenkel, L.
\newblock Tree-based batch mode reinforcement learning.
\newblock \emph{Journal of Machine Learning Research}, 6\penalty0
  (Apr):\penalty0 503--556, 2005.

\bibitem[Farahmand \& Szepesv{\'a}ri(2011)Farahmand and
  Szepesv{\'a}ri]{farahmand2011model}
Farahmand, A.-m. and Szepesv{\'a}ri, C.
\newblock Model selection in reinforcement learning.
\newblock \emph{Machine learning}, 85\penalty0 (3):\penalty0 299--332, 2011.

\bibitem[Farahmand et~al.(2010)Farahmand, Munos, and
  Szepesv{\'a}ri]{farahmand2010error}
Farahmand, A.~M., Munos, R., and Szepesv{\'a}ri, C.
\newblock Error propagation for approximate policy and value iteration.
\newblock In \emph{Advances in Neural Information Processing Systems}, 2010.

\bibitem[Feng et~al.(2019)Feng, Li, and Liu]{feng2019kernel}
Feng, Y., Li, L., and Liu, Q.
\newblock A kernel loss for solving the bellman equation.
\newblock \emph{Advances in Neural Information Processing Systems},
  32:\penalty0 15456--15467, 2019.

\bibitem[Fu et~al.(2019)Fu, Kumar, Soh, and Levine]{fu2019diagnosing}
Fu, J., Kumar, A., Soh, M., and Levine, S.
\newblock Diagnosing bottlenecks in deep q-learning algorithms.
\newblock In \emph{International Conference on Machine Learning}, pp.\
  2021--2030. PMLR, 2019.

\bibitem[Fu et~al.(2021)Fu, Norouzi, Nachum, Tucker, Wang, Novikov, Yang,
  Zhang, Chen, Kumar, Paduraru, Levine, and Paine]{fu2021benchmarks}
Fu, J., Norouzi, M., Nachum, O., Tucker, G., Wang, Z., Novikov, A., Yang, M.,
  Zhang, M.~R., Chen, Y., Kumar, A., Paduraru, C., Levine, S., and Paine, T.
\newblock Benchmarks for deep off-policy evaluation.
\newblock In \emph{International Conference on Learning Representations}, 2021.

\bibitem[Fujimoto et~al.(2018)Fujimoto, van Hoof, and
  Meger]{fujimoto2018addressing}
Fujimoto, S., van Hoof, H., and Meger, D.
\newblock Addressing function approximation error in actor-critic methods.
\newblock In \emph{International Conference on Machine Learning}, volume~80,
  pp.\  1587--1596. PMLR, 2018.

\bibitem[Fujimoto et~al.(2019{\natexlab{a}})Fujimoto, Conti, Ghavamzadeh, and
  Pineau]{fujimoto2019benchmarking}
Fujimoto, S., Conti, E., Ghavamzadeh, M., and Pineau, J.
\newblock Benchmarking batch deep reinforcement learning algorithms.
\newblock \emph{arXiv preprint arXiv:1910.01708}, 2019{\natexlab{a}}.

\bibitem[Fujimoto et~al.(2019{\natexlab{b}})Fujimoto, Meger, and
  Precup]{fujimoto2018off}
Fujimoto, S., Meger, D., and Precup, D.
\newblock Off-policy deep reinforcement learning without exploration.
\newblock In \emph{International Conference on Machine Learning}, pp.\
  2052--2062, 2019{\natexlab{b}}.

\bibitem[Fujimoto et~al.(2021)Fujimoto, Meger, and Precup]{fujimoto2021srdice}
Fujimoto, S., Meger, D., and Precup, D.
\newblock A deep reinforcement learning approach to marginalized importance
  sampling with the successor representation.
\newblock In \emph{Proceedings of the 38th International Conference on Machine
  Learning}, volume 139, pp.\  3518--3529. PMLR, 2021.

\bibitem[Geist et~al.(2017)Geist, Piot, and Pietquin]{geist2017bellman}
Geist, M., Piot, B., and Pietquin, O.
\newblock Is the bellman residual a bad proxy?
\newblock In \emph{Proceedings of the 31st International Conference on Neural
  Information Processing Systems}, pp.\  3208--3217, 2017.

\bibitem[Gu et~al.(2016)Gu, Lillicrap, Sutskever, and Levine]{gu2016continuous}
Gu, S., Lillicrap, T., Sutskever, I., and Levine, S.
\newblock Continuous deep q-learning with model-based acceleration.
\newblock In \emph{International Conference on Machine Learning}, pp.\
  2829--2838, 2016.

\bibitem[Haarnoja et~al.(2018{\natexlab{a}})Haarnoja, Zhou, Abbeel, and
  Levine]{haarnoja2018soft}
Haarnoja, T., Zhou, A., Abbeel, P., and Levine, S.
\newblock Soft actor-critic: Off-policy maximum entropy deep reinforcement
  learning with a stochastic actor.
\newblock In \emph{International Conference on Machine Learning}, volume~80,
  pp.\  1861--1870. PMLR, 2018{\natexlab{a}}.

\bibitem[Haarnoja et~al.(2018{\natexlab{b}})Haarnoja, Zhou, Hartikainen,
  Tucker, Ha, Tan, Kumar, Zhu, Gupta, Abbeel, et~al.]{haarnoja2018applications}
Haarnoja, T., Zhou, A., Hartikainen, K., Tucker, G., Ha, S., Tan, J., Kumar,
  V., Zhu, H., Gupta, A., Abbeel, P., et~al.
\newblock Soft actor-critic algorithms and applications.
\newblock \emph{arXiv preprint arXiv:1812.05905}, 2018{\natexlab{b}}.

\bibitem[Heger(1996)]{heger1996loss}
Heger, M.
\newblock The loss from imperfect value functions in expectation-based and
  minimax-based tasks.
\newblock \emph{Machine Learning}, 22\penalty0 (1):\penalty0 197--225, 1996.

\bibitem[Henderson et~al.(2017)Henderson, Islam, Bachman, Pineau, Precup, and
  Meger]{henderson2017deep}
Henderson, P., Islam, R., Bachman, P., Pineau, J., Precup, D., and Meger, D.
\newblock Deep reinforcement learning that matters.
\newblock In \emph{AAAI Conference on Artificial Intelligence}, 2017.

\bibitem[Hessel et~al.(2018)Hessel, Modayil, Van~Hasselt, Schaul, Ostrovski,
  Dabney, Horgan, Piot, Azar, and Silver]{hessel2018rainbow}
Hessel, M., Modayil, J., Van~Hasselt, H., Schaul, T., Ostrovski, G., Dabney,
  W., Horgan, D., Piot, B., Azar, M., and Silver, D.
\newblock Rainbow: Combining improvements in deep reinforcement learning.
\newblock In \emph{Thirty-second AAAI conference on artificial intelligence},
  2018.

\bibitem[Irpan et~al.(2019)Irpan, Rao, Bousmalis, Harris, Ibarz, and
  Levine]{irpan2019off}
Irpan, A., Rao, K., Bousmalis, K., Harris, C., Ibarz, J., and Levine, S.
\newblock Off-policy evaluation via off-policy classification.
\newblock \emph{Advances in Neural Information Processing Systems},
  32:\penalty0 5437--5448, 2019.

\bibitem[Jaderberg et~al.(2016)Jaderberg, Mnih, Czarnecki, Schaul, Leibo,
  Silver, and Kavukcuoglu]{jaderberg2016reinforcement}
Jaderberg, M., Mnih, V., Czarnecki, W.~M., Schaul, T., Leibo, J.~Z., Silver,
  D., and Kavukcuoglu, K.
\newblock Reinforcement learning with unsupervised auxiliary tasks.
\newblock \emph{arXiv preprint arXiv:1611.05397}, 2016.

\bibitem[Kakade \& Langford(2002)Kakade and Langford]{kakade2002approximately}
Kakade, S. and Langford, J.
\newblock Approximately optimal approximate reinforcement learning.
\newblock In \emph{International Conference on Machine Learning}, volume~2,
  pp.\  267--274, 2002.

\bibitem[Kingma \& Ba(2014)Kingma and Ba]{adam}
Kingma, D. and Ba, J.
\newblock Adam: A method for stochastic optimization.
\newblock \emph{arXiv preprint arXiv:1412.6980}, 2014.

\bibitem[Kolter(2011)]{kolter2011fixed}
Kolter, J.~Z.
\newblock The fixed points of off-policy td.
\newblock In \emph{Advances in Neural Information Processing Systems}, 2011.

\bibitem[Lange et~al.(2012)Lange, Gabel, and Riedmiller]{lange2012batch}
Lange, S., Gabel, T., and Riedmiller, M.
\newblock Batch reinforcement learning.
\newblock In \emph{Reinforcement learning}, pp.\  45--73. Springer, 2012.

\bibitem[Le et~al.(2019)Le, Voloshin, and Yue]{le2019batch}
Le, H., Voloshin, C., and Yue, Y.
\newblock Batch policy learning under constraints.
\newblock In \emph{International Conference on Machine Learning}, pp.\
  3703--3712. PMLR, 2019.

\bibitem[Levine et~al.(2020)Levine, Kumar, Tucker, and Fu]{levine2020offline}
Levine, S., Kumar, A., Tucker, G., and Fu, J.
\newblock Offline reinforcement learning: Tutorial, review, and perspectives on
  open problems.
\newblock \emph{arXiv preprint arXiv:2005.01643}, 2020.

\bibitem[Lillicrap et~al.(2015)Lillicrap, Hunt, Pritzel, Heess, Erez, Tassa,
  Silver, and Wierstra]{DDPG}
Lillicrap, T.~P., Hunt, J.~J., Pritzel, A., Heess, N., Erez, T., Tassa, Y.,
  Silver, D., and Wierstra, D.
\newblock Continuous control with deep reinforcement learning.
\newblock \emph{arXiv preprint arXiv:1509.02971}, 2015.

\bibitem[Mnih et~al.(2015)Mnih, Kavukcuoglu, Silver, Rusu, Veness, Bellemare,
  Graves, Riedmiller, Fidjeland, Ostrovski, et~al.]{DQN}
Mnih, V., Kavukcuoglu, K., Silver, D., Rusu, A.~A., Veness, J., Bellemare,
  M.~G., Graves, A., Riedmiller, M., Fidjeland, A.~K., Ostrovski, G., et~al.
\newblock Human-level control through deep reinforcement learning.
\newblock \emph{Nature}, 518\penalty0 (7540):\penalty0 529--533, 2015.

\bibitem[Munos(2003)]{munos2003error}
Munos, R.
\newblock Error bounds for approximate policy iteration.
\newblock In \emph{ICML}, volume~3, pp.\  560--567, 2003.

\bibitem[Munos(2007)]{munos2007performance}
Munos, R.
\newblock Performance bounds in l\_p-norm for approximate value iteration.
\newblock \emph{SIAM journal on control and optimization}, 46\penalty0
  (2):\penalty0 541--561, 2007.

\bibitem[Paine et~al.(2020)Paine, Paduraru, Michi, Gulcehre, Zolna, Novikov,
  Wang, and de~Freitas]{paine2020hyperparameter}
Paine, T.~L., Paduraru, C., Michi, A., Gulcehre, C., Zolna, K., Novikov, A.,
  Wang, Z., and de~Freitas, N.
\newblock Hyperparameter selection for offline reinforcement learning.
\newblock \emph{arXiv preprint arXiv:2007.09055}, 2020.

\bibitem[Paszke et~al.(2019)Paszke, Gross, Massa, Lerer, Bradbury, Chanan,
  Killeen, Lin, Gimelshein, Antiga, et~al.]{paszke2019pytorch}
Paszke, A., Gross, S., Massa, F., Lerer, A., Bradbury, J., Chanan, G., Killeen,
  T., Lin, Z., Gimelshein, N., Antiga, L., et~al.
\newblock Pytorch: An imperative style, high-performance deep learning library.
\newblock In \emph{Advances in Neural Information Processing Systems}, pp.\
  8024--8035, 2019.

\bibitem[Patterson et~al.(2022)Patterson, White, and
  White]{patterson2022generalized}
Patterson, A., White, A., and White, M.
\newblock A generalized projected bellman error for off-policy value estimation
  in reinforcement learning.
\newblock \emph{Journal of Machine Learning Research}, 23\penalty0
  (145):\penalty0 1--61, 2022.

\bibitem[Puterman(1994)]{puterman1994markov}
Puterman, M.~L.
\newblock \emph{Markov Decision Processes: Discrete Stochastic Dynamic
  Programming}.
\newblock John Wiley \& Sons, Inc., 1994.

\bibitem[Saleh \& Jiang(2019)Saleh and Jiang]{saleh2019deterministic}
Saleh, E. and Jiang, N.
\newblock Deterministic bellman residual minimization.
\newblock In \emph{Proceedings of Optimization Foundations for Reinforcement
  Learning Workshop at NeurIPS}, 2019.

\bibitem[Schulman et~al.(2017)Schulman, Wolski, Dhariwal, Radford, and
  Klimov]{ppo}
Schulman, J., Wolski, F., Dhariwal, P., Radford, A., and Klimov, O.
\newblock Proximal policy optimization algorithms.
\newblock \emph{arXiv preprint arXiv:1707.06347}, 2017.

\bibitem[Schweitzer \& Seidmann(1985)Schweitzer and
  Seidmann]{schweitzer1985generalized}
Schweitzer, P.~J. and Seidmann, A.
\newblock Generalized polynomial approximations in markovian decision
  processes.
\newblock \emph{Journal of mathematical analysis and applications},
  110\penalty0 (2):\penalty0 568--582, 1985.

\bibitem[Singh \& Yee(1994)Singh and Yee]{singh1994upper}
Singh, S.~P. and Yee, R.~C.
\newblock An upper bound on the loss from approximate optimal-value functions.
\newblock \emph{Machine Learning}, 16\penalty0 (3):\penalty0 227--233, 1994.

\bibitem[Sutton(1988)]{sutton1988tdlearning}
Sutton, R.~S.
\newblock Learning to predict by the methods of temporal differences.
\newblock \emph{Machine learning}, 3\penalty0 (1):\penalty0 9--44, 1988.

\bibitem[Sutton \& Barto(1998)Sutton and Barto]{sutton1998reinforcement}
Sutton, R.~S. and Barto, A.~G.
\newblock \emph{Reinforcement learning: An introduction}, volume~1.
\newblock MIT press Cambridge, 1998.

\bibitem[Sutton et~al.(2009)Sutton, Maei, Precup, Bhatnagar, Silver,
  Szepesv{\'a}ri, and Wiewiora]{sutton2009fast}
Sutton, R.~S., Maei, H.~R., Precup, D., Bhatnagar, S., Silver, D.,
  Szepesv{\'a}ri, C., and Wiewiora, E.
\newblock Fast gradient-descent methods for temporal-difference learning with
  linear function approximation.
\newblock In \emph{Proceedings of the 26th Annual International Conference on
  Machine Learning}, pp.\  993--1000, 2009.

\bibitem[Tang \& Wiens(2021)Tang and Wiens]{tang2021model}
Tang, S. and Wiens, J.
\newblock Model selection for offline reinforcement learning: Practical
  considerations for healthcare settings.
\newblock In \emph{Machine Learning for Healthcare Conference}, pp.\  2--35.
  PMLR, 2021.

\bibitem[Todorov et~al.(2012)Todorov, Erez, and Tassa]{mujoco}
Todorov, E., Erez, T., and Tassa, Y.
\newblock Mujoco: A physics engine for model-based control.
\newblock In \emph{IEEE/RSJ International Conference on Intelligent Robots and
  Systems (IROS)}, pp.\  5026--5033. IEEE, 2012.

\bibitem[Voloshin et~al.(2021)Voloshin, Le, Jiang, and
  Yue]{voloshin2021empirical}
Voloshin, C., Le, H.~M., Jiang, N., and Yue, Y.
\newblock Empirical study of off-policy policy evaluation for reinforcement
  learning.
\newblock In \emph{Thirty-fifth Conference on Neural Information Processing
  Systems Datasets and Benchmarks Track (Round 1)}, 2021.

\bibitem[Wang et~al.(2020)Wang, Foster, and Kakade]{wang2020statistical}
Wang, R., Foster, D., and Kakade, S.~M.
\newblock What are the statistical limits of offline rl with linear function
  approximation?
\newblock In \emph{International Conference on Learning Representations}, 2020.

\bibitem[Williams \& Baird(1993)Williams and Baird]{williams1993tight}
Williams, R.~J. and Baird, L.
\newblock Tight performance bounds on greedy policies based on imperfect value
  functions.
\newblock Technical report, Northeastern University, College of Computer
  Science, 1993.

\bibitem[Xiao et~al.(2021)Xiao, Lee, Dai, Schuurmans, and
  Szepesvari]{xiao2021sample}
Xiao, C., Lee, I., Dai, B., Schuurmans, D., and Szepesvari, C.
\newblock On the sample complexity of batch reinforcement learning with
  policy-induced data.
\newblock \emph{arXiv preprint arXiv:2106.09973}, 2021.

\bibitem[Yang \& Nachum(2021)Yang and Nachum]{yang2021representation}
Yang, M. and Nachum, O.
\newblock Representation matters: Offline pretraining for sequential decision
  making.
\newblock In \emph{Self-Supervision for Reinforcement Learning Workshop-ICLR
  2021}, 2021.

\bibitem[Zanette(2021)]{zanette2021exponential}
Zanette, A.
\newblock Exponential lower bounds for batch reinforcement learning: Batch rl
  can be exponentially harder than online rl.
\newblock In \emph{International Conference on Machine Learning}, pp.\
  12287--12297. PMLR, 2021.

\bibitem[Zhang et~al.(2020)Zhang, Boehmer, and Whiteson]{zhang2020deep}
Zhang, S., Boehmer, W., and Whiteson, S.
\newblock Deep residual reinforcement learning.
\newblock In \emph{Proceedings of the 19th International Conference on
  Autonomous Agents and MultiAgent Systems}, pp.\  1611--1619, 2020.

\bibitem[Zhu \& Ying(2020)Zhu and Ying]{zhu2020borrowing}
Zhu, Y. and Ying, L.
\newblock Borrowing from the future: An attempt to address double sampling.
\newblock In \emph{Mathematical and scientific machine learning}, pp.\
  246--268. PMLR, 2020.

\end{thebibliography}
\bibliographystyle{icml2022}

\clearpage

\onecolumn

\setcounter{theorem}{0}
\setcounter{corollary}{0}
\setcounter{observation}{0}
\setcounter{lemma}{0}
\setcounter{proposition}{0}

\appendix

\section{Additional Figures}\label{appendix:final}

\subsection{Is the Bellman error a valid on-policy objective?}\label{appendix:correlation}

\begin{table*}[ht]
\small
\centering
\begin{tabular}{ccccccccc}
\toprule
Train Data & Test Data & Algorithm & \makebox[4em][c]{HalfCheetah} & \makebox[4em][c]{Hopper} & \makebox[4em][c]{Walker2d} & \makebox[4em][c]{Ant} & \makebox[4em][c]{Humanoid} \\ 
\midrule
\multirow{2}{*}{All} & \multirow{2}{*}{On-Policy} & BRM  & \cblockhtml{E04419}~\hphantom{-}0.95  & \cblockhtml{E66D4D}~\hphantom{-}0.74  & \cblockhtml{E04116}~\hphantom{-}0.96  & \cblockhtml{DF3C10}~\hphantom{-}0.99  & \cblockhtml{DF3D11}~\hphantom{-}0.98  \\
& & FQE   & \cblockhtml{E45F3B}~\hphantom{-}0.81  & \cblockhtml{E66947}~\hphantom{-}0.76  & \cblockhtml{E77051}~\hphantom{-}0.72  & \cblockhtml{E56341}~\hphantom{-}0.79  & \cblockhtml{F9E6E4}~\hphantom{-}0.11  \\
\midrule %
\multirow{2}{*}{All} & \multirow{2}{*}{$0.1$} & BRM     & \cblockhtml{B0C7E4}~-0.46 & \cblockhtml{769DCF}~-0.83 & \cblockhtml{83A7D4}~-0.74 & \cblockhtml{82A6D4}~-0.75 & \cblockhtml{93B2DA}~-0.65 \\
& & FQE      & \cblockhtml{EB8C74}~\hphantom{-}0.57  & \cblockhtml{E35732}~\hphantom{-}0.85  & \cblockhtml{6A95CB}~-0.90 & \cblockhtml{9AB7DC}~-0.60 & \cblockhtml{F6D5CE}~\hphantom{-}0.20  \\
\midrule
\multirow{2}{*}{$0.1$} & \multirow{2}{*}{$0.1$} & BRM & \cblockhtml{F9E6E4}~\hphantom{-}0.11 & \cblockhtml{FBF4F5}~\hphantom{-}0.04 & \cblockhtml{AFC5E3}~-0.47 & \cblockhtml{EFA28F}~\hphantom{-}0.46 & \cblockhtml{ADC4E3}~-0.48 \\
& & FQE & \cblockhtml{E14A21}~\hphantom{-}0.92 & \cblockhtml{F4C4BA}~\hphantom{-}0.29 & \cblockhtml{E5ECF7}~-0.14 & \cblockhtml{EB8C74}~\hphantom{-}0.58 & \cblockhtml{FBF2F2}~\hphantom{-}0.05  \\
\bottomrule
\end{tabular}
\caption[]{\textbf{Is there a correlation between Bellman error and value error?} We display Pearson's correlation coefficient of the final Bellman error and value error of functions trained with either only BRM or only FQE. Warm colors\hspace{-0.15em}\cblockhtml{E04419} are used to show positive correlation and cold colors\hspace{-0.15em}\cblockhtml{769DCF} are used for negative correlation. The error terms are computed over held-out rollouts. The functions are trained using datasets of varying noise levels, where all refers to the set ($0.1$, $0.2$, $0.3$, $0.4$, $0.5$) with 10 seeds, (6$\times$10 functions), 
$0.1$ refers to the subset of functions trained on the $0.1$ dataset (10 functions). 
While there is high correlation between the on-policy empirical Bellman error and value error when comparing functions trained with the same algorithm, this relationship is not strong when evaluated with an off-policy dataset $(0.1)$.} \label{table:correlation} %
\end{table*}

\subsection{Does FQE provide a better metric?}\label{appendix:correlation_FQE}

\begin{table*}[ht]
\small
\centering
\begin{tabular}{ccccccccc}
\toprule
Train Data & Test Data & Metric & \makebox[4em][c]{HalfCheetah} & \makebox[4em][c]{Hopper} & \makebox[4em][c]{Walker2d} & \makebox[4em][c]{Ant} & \makebox[4em][c]{Humanoid} \\ 
\midrule
\multirow{2}{*}{All} & \multirow{2}{*}{On-Policy} & BE & \cblockhtml{E45F3B}~\hphantom{-}0.81  & \cblockhtml{E66947}~\hphantom{-}0.76  & \cblockhtml{E77051}~\hphantom{-}0.72  & \cblockhtml{E56341}~\hphantom{-}0.79  & \cblockhtml{F9E6E4}~\hphantom{-}0.11  \\
& & $\Loss_\text{FQE}$ & \cblockhtml{E45F3B}~\hphantom{-}0.81 & \cblockhtml{E56340}~\hphantom{-}0.79 & \cblockhtml{EA876D}~\hphantom{-}0.60  & \cblockhtml{E45E3A}~\hphantom{-}0.81  & \cblockhtml{F6D1CA}~\hphantom{-}0.22 \\
\midrule %
\multirow{2}{*}{All} & \multirow{2}{*}{$0.1$} & BE & \cblockhtml{EB8C74}~\hphantom{-}0.57  & \cblockhtml{E35732}~\hphantom{-}0.85  & \cblockhtml{6A95CB}~-0.90 & \cblockhtml{9AB7DC}~-0.60 & \cblockhtml{F6D5CE}~\hphantom{-}0.20  \\
& & $\Loss_\text{FQE}$ & \cblockhtml{EA8469}~\hphantom{-}0.62 & \cblockhtml{E35833}~\hphantom{-}0.84 & \cblockhtml{6994CB}~-0.90 & \cblockhtml{9CB8DD}~-0.59 & \cblockhtml{F6D3CC}~\hphantom{-}0.21 \\
\bottomrule
\end{tabular}
\caption[]{\textbf{Is there a correlation between the FQE loss and value error?} Pearson's correlation coefficient of the Bellman error (BE) and the FQE objective ($\Loss_\text{FQE}$) with the value error of functions trained by FQE.  
Warm colors\hspace{-0.15em}\cblockhtml{E04419} are used to show positive correlation and cold colors\hspace{-0.15em}\cblockhtml{769DCF} are used for negative correlation. The error terms are computed over held-out rollouts. The functions are trained using datasets of varying noise levels, where all refers to the set ($0.1$, $0.2$, $0.3$, $0.4$, $0.5$) with 10 seeds, (6$\times$10 functions), 
$0.1$ refers to the subset of functions trained on the $0.1$ dataset (10 functions). We can see that the difference between BE and $\Loss_\text{FQE}$ is minimal. This means that the FQE loss function is not a better metric than BE. This is logical as it is non-stationary and dependent on the current value function. 
} \label{table:correlation_FQE}
\end{table*}

\subsection{Additional Single Trajectory Experiments} \label{appendix:one_traj}

\begin{figure}[ht]
    \centering
    
    \subfloat[HalfCheetah]{
    \includegraphics[width=0.24\linewidth]{images/one_traj_bellman.png}
    \includegraphics[width=0.24\linewidth]{images/one_traj_value.png}
    }
    \subfloat[Hopper]{
    \includegraphics[width=0.24\linewidth]{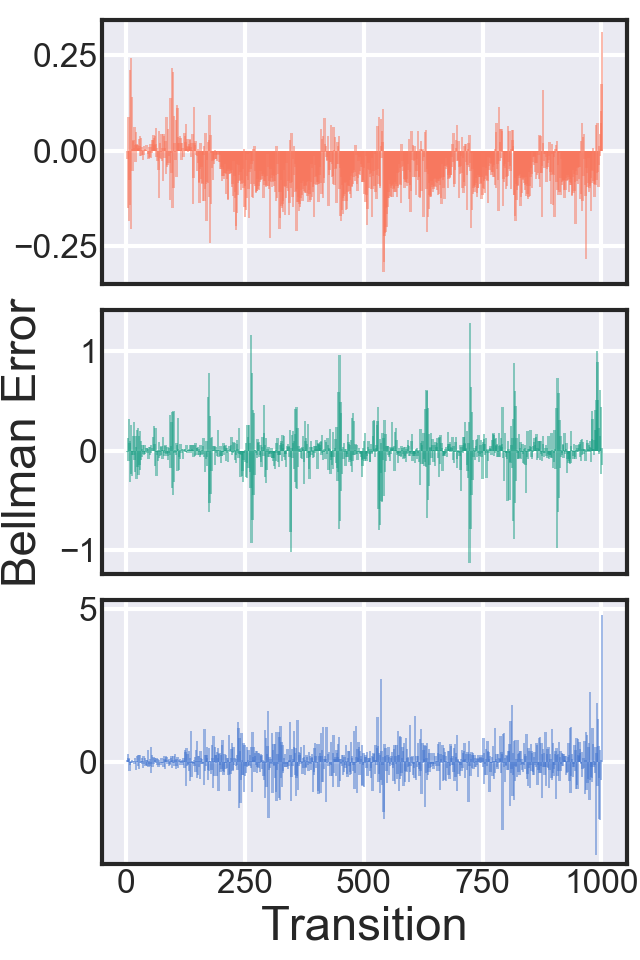}
    \includegraphics[width=0.24\linewidth]{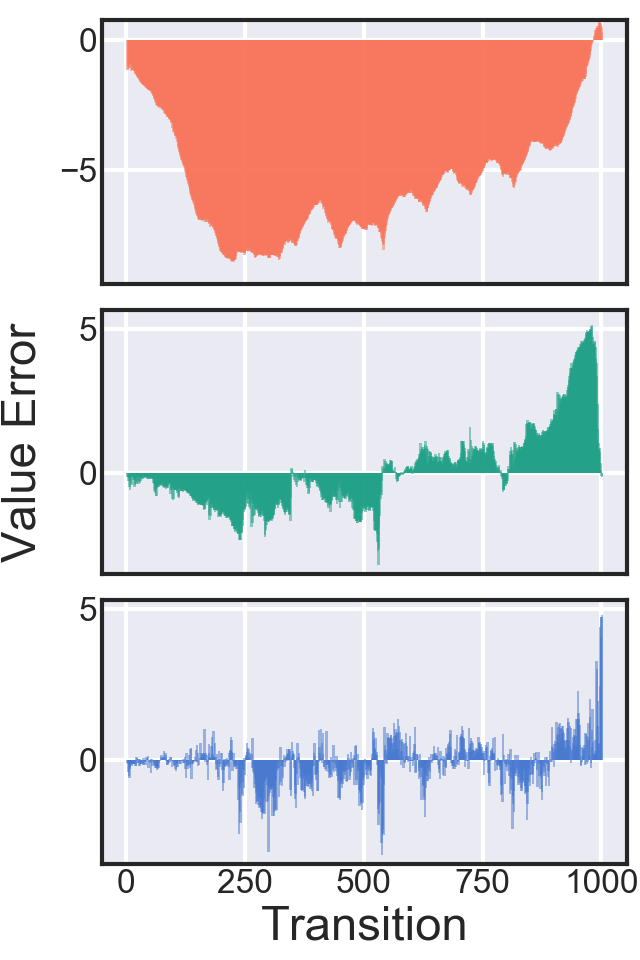}
    }
    
    \subfloat[Walker2d]{
    \includegraphics[width=0.24\linewidth]{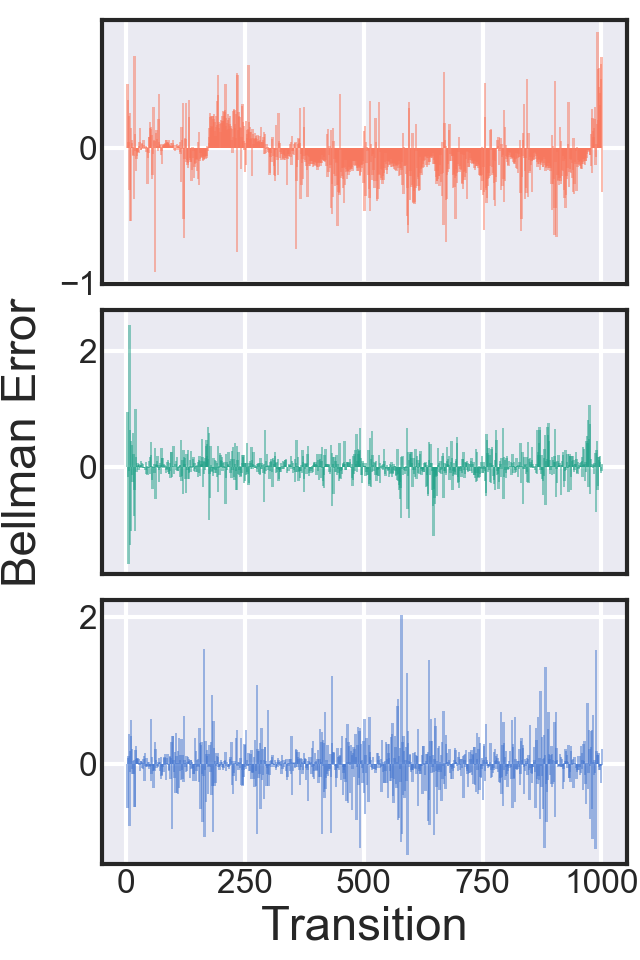}
    \includegraphics[width=0.24\linewidth]{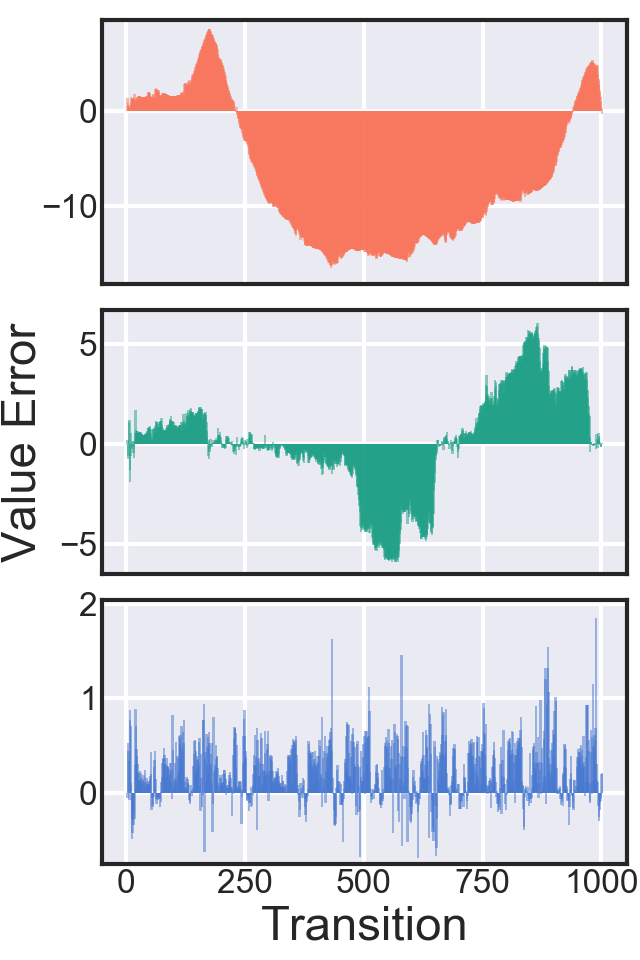}
    }
    \subfloat[Ant]{
    \includegraphics[width=0.24\linewidth]{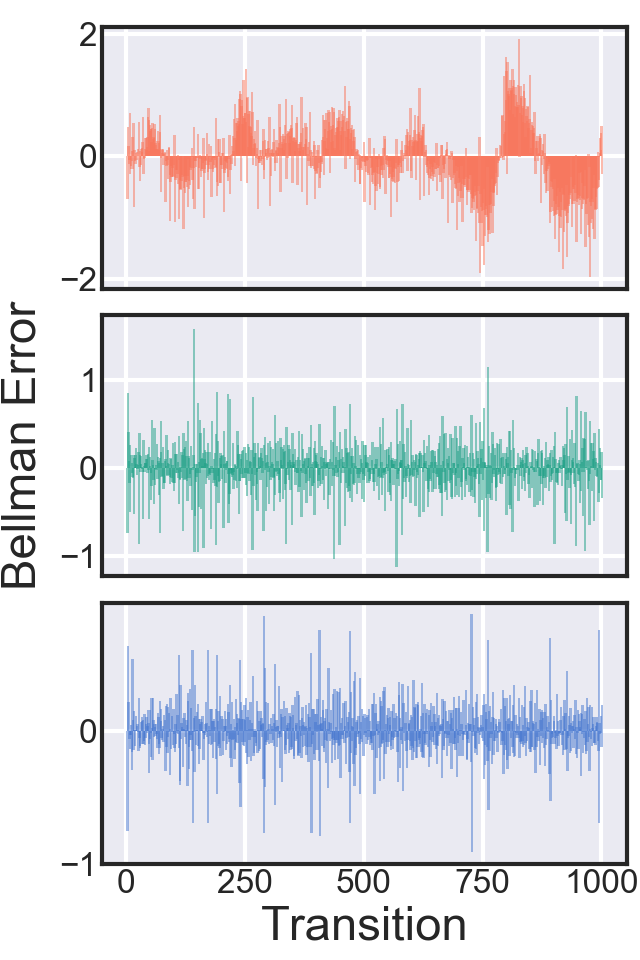}
    \includegraphics[width=0.24\linewidth]{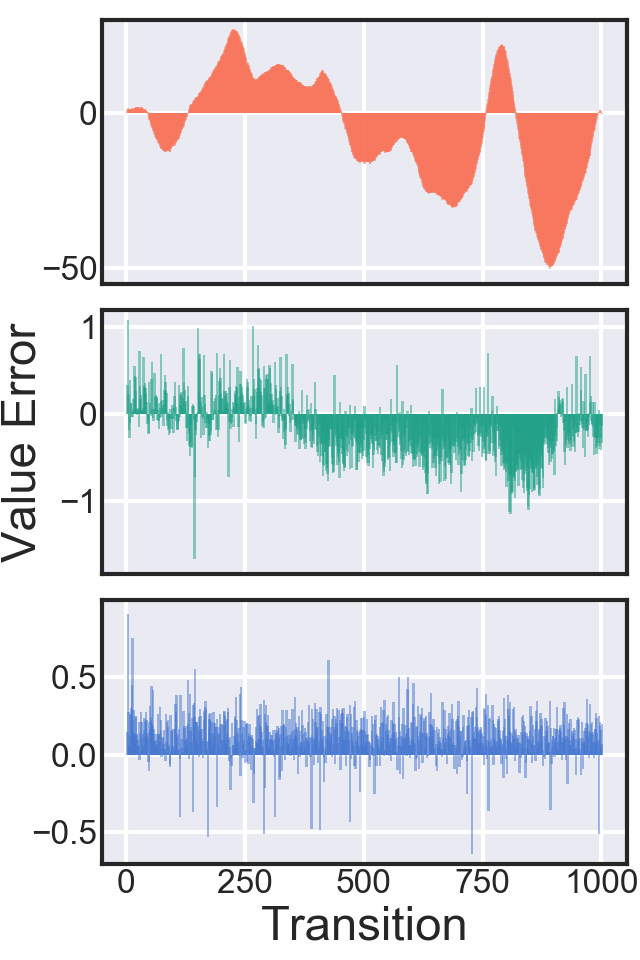}
    }
    
    \subfloat[Humanoid]{
    \includegraphics[width=0.24\linewidth]{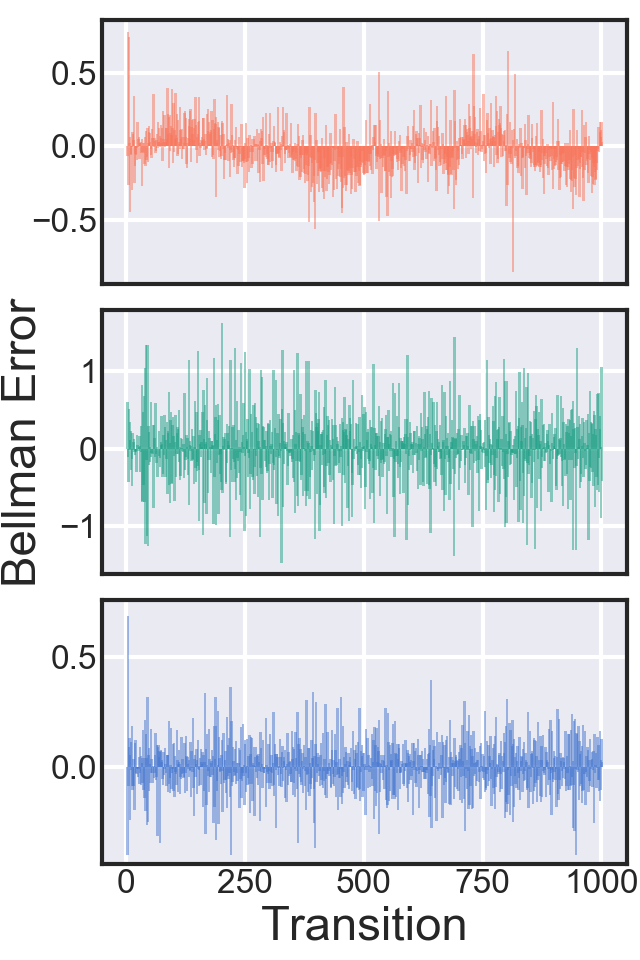}
    \includegraphics[width=0.24\linewidth]{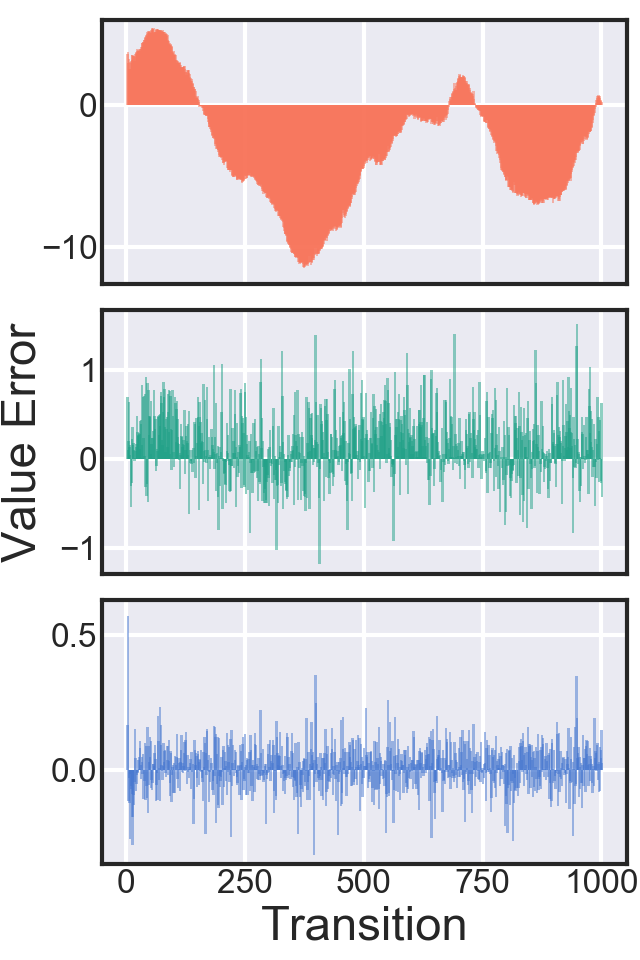}
    }
    
    \vspace{4.5pt}
    \small \cblock{my_warm_2} BRM \qquad \cblock{my_cold_3} FQE \qquad \cblock{my_blue} MC

\caption{\textbf{Does bias explain the non-correspondence between the Bellman error and value error?} We train value functions over a single on-policy trajectory (1000 transitions, with termination). As a result, every relevant transition is contained in each dataset. We display the Bellman error (left) and unnormalized value error (right) for every transition in the trajectory. Visually, we can see that the clusters of similar value error (size and direction) can cause lower magnitude Bellman error for\cblock{my_warm_2}~BRM and\cblock{my_cold_3}~FQE.}
\end{figure}

\begin{table}[h]
\small
\centering
\setlength{\tabcolsep}{10pt}
\begin{tabular}{lrrrrrrrrr}
\toprule
            & \multicolumn{3}{c}{BRM} & \multicolumn{3}{c}{FQE} & \multicolumn{3}{c}{MC} \\ \midrule
            & BE    & VE     & Ratio  & BE     & VE    & Ratio  & BE    & VE    & Ratio  \\ \midrule
HalfCheetah & 0.41  & 23.39  & 56.93  & 0.76   & 8.89  & 11.72  & 2.76  & 2.75  & 1.00   \\
Hopper      & 0.07  & 5.54   & 77.43  & 0.12   & 1.13  & 9.33   & 0.36  & 0.50  & 1.39   \\
Walker2d    & 0.16  & 8.87   & 56.56  & 0.16   & 1.92  & 11.79  & 0.19  & 0.29  & 1.50   \\
Ant         & 0.42  & 16.02  & 38.17  & 0.20   & 0.33  & 1.68   & 0.14  & 0.15  & 1.09   \\
Humanoid    & 0.12  & 4.50   & 37.64  & 0.33   & 0.30  & 0.93   & 0.09  & 0.06  & 0.72   \\ \bottomrule
\end{tabular}
\caption{\textbf{Does bias explain the non-correspondence between the Bellman error and value error?} We train value functions over a single on-policy trajectory (1000 transitions, with termination). As a result, every relevant transition is contained in each dataset. We report the average absolute Bellman error (BE) and value error (VE), and ratio between these terms. By \autoref{prop:bounds}, with a discount $\y=0.99$, the ratio between mean absolute Bellman and value error is bounded by $(0.5, 100)$. The tabular results show that both the upper and lower bound can be approached in practice.} 
\end{table}

\clearpage

\subsection{BRM Behaves Predictably} \label{appendix:BRM_predict}

\begin{figure}[ht]
    \centering
    \includegraphics[width=\textwidth]{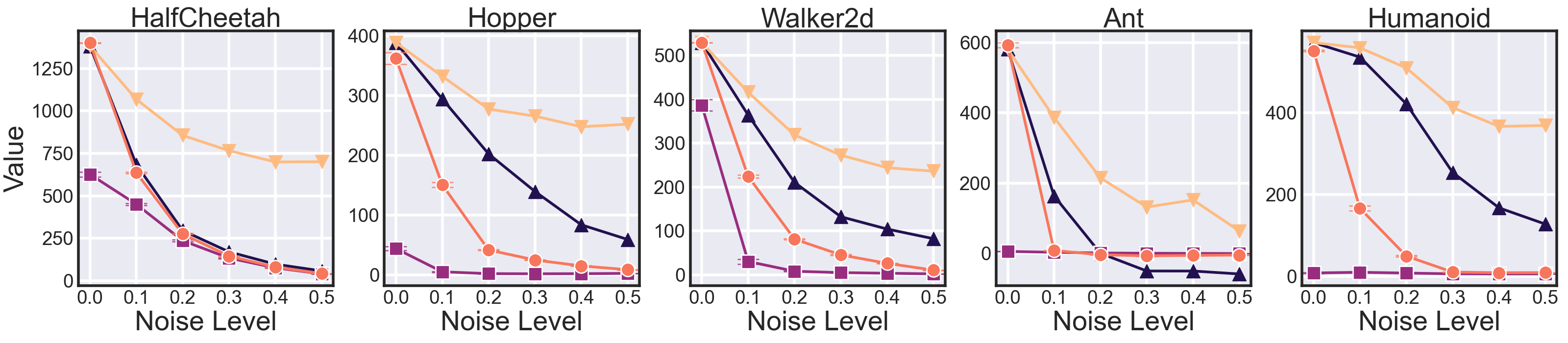}
    \small \ctriangle{my_warm_1}Target Policy \qquad \cinvtriangle{my_warm_6}Behavior Policy \qquad \ccircle{my_warm_2}BRM \qquad \cblock{my_warm_4} BRM (Suboptimal target policy)
    \caption[]{Visualizing the final value estimated by BRM after training on different datasets corresponding to varying noise levels. The true value of the target policy and the behavior policy are displayed to provide reference, as well as BRM when trained to evaluate a suboptimal policy (corresponding to TD3 trained for 300k time steps rather than 3m). Error bars capture the standard deviation over 10 seeds (but are visually hard to see as the deviation is low). 
    We can see that\ccircle{my_warm_2}BRM typically converges to a value which is closer to the\cinvtriangle{my_warm_6}behavior policy rather than the\ctriangle{my_warm_1}target policy, and typically prefers values which are close to 0. Interestingly, the BRM trained to evaluate the suboptimal target policy\cblock{my_warm_4} converges to the same value on the noisiest datasets, suggesting that the influence of the target policy on what value BRM converges to is reduced with increased distribution shift.} \label{fig:convergence}
\end{figure}

\clearpage

\section{Proofs}

\subsection{Proposition 1} %

\begin{proposition}
The Bellman error $\e(s,a)=0$ for all state-action pairs $(s,a) \in \mathcal{S} \times \mathcal{A}$, if and only if the value error $\Delta(s,a)=0$ for all state-action pairs $(s,a) \in \mathcal{S} \times \mathcal{A}$.
\end{proposition}

\textit{Proof.}

$\Rightarrow$: This is a direct consequence of \autoref{appendix:thm1}.

$\Leftarrow$: If for all $(s,a) \in \mathcal{S} \times \mathcal{A}$, $\Delta(s,a) = Q(s,a) - Q^\pi(s,a) = 0$ then $Q(s,a) = Q^\pi(s,a)$ for all $(s,a) \in \mathcal{S} \times \mathcal{A}$. From the Bellman equation it follows that 
\begin{align}
    Q(s,a) &= Q^\pi(s,a) \\
    &= \mathcal{T} Q^\pi(s,a) \\
    &= \mathcal{T} Q(s,a).
\end{align}

\hfill $\blacksquare$

\subsection{Proposition 2}

\begin{proposition}
Let $Q_1, Q_2$ be value functions, where $Q_2 = \mathcal{T} Q_1$. $\max_{(s,a)} |\Delta_{Q_2}(s,a)| \leq \y \cdot \max_{(s,a)} |\Delta_{Q_1}(s,a)|$. 
\end{proposition}

\textit{Proof.}

For all $(s,a) \in \mathcal{S} \times \mathcal{A}$: 
\begin{align}
    \Delta_2(s,a) &= \mathcal{T} Q_1(s,a) - Q^\pi(s,a) \\
    &= \mathcal{T} Q_1(s,a) - \mathcal{T} Q^\pi(s,a) \\
    &= \E_{s',a'}[\y (Q_1(s',a') - Q^\pi(s',a'))] \\
    &= \E_{s',a'}[\y \Delta_1(s',a')].
\end{align}
It follows that for all $(s,a) \in \mathcal{S} \times \mathcal{A}$: 
\begin{align}
    |\Delta_2(s,a)| &\leq \max_{(s,a)} \y |\Delta_1(s,a)| \\
    \Rightarrow \max_{(s,a)} |\Delta_2(s,a)| &\leq \max_{(s,a)} \y |\Delta_1(s,a)|.
\end{align}

\hfill $\blacksquare$

\subsection{Proposition 3}

\begin{proposition} {\normalfont \textbf{(An inverse relationship between Bellman error and value error).}}
For any MDP, discount factor $\y \in (0,1)$, and $C>0$, we can define a value function $Q_1$ and a stochastic value function $Q_2$ such that for any state-action pair~$(s,a)\in \mathcal{S} \times \mathcal{A}$
\begin{enumerate}[nosep]
    \item $|\Delta_{Q_1}(s,a)| - |\Delta_{Q_2}(s,a)| > C$,
    \item $\E_{Q_2}[|\e_{Q_2}(s,a)|] - |\e_{Q_1}(s,a)| > C$.
\end{enumerate}
\end{proposition}

\textit{Proof.}

Let
\begin{align}
    Q_1(s,a) &= Q^\pi(s,a) + \frac{k}{1-\y}, \\
    Q_2(s,a) &= Q^\pi(s,a) \pm k(1+\y).
\end{align}

Then
\begin{align}
|\Delta_1(s,a)| &= \frac{k}{1-\y}, \\
|\Delta_2(s,a)| &= k(1+\y),
\end{align}
where $\frac{k}{1-\y} > k(1+\y)$ for all $\y \in (0,1)$.

And
\begin{align}
\e_1(s,a) &= \frac{k}{1-\y} - \frac{\y k}{1-\y} = k, \\
\E[|\e_2(s,a)|] &= 0.5 | k(1+\y) - \y k(1+\y) |  + 0.5 | k(1+\y) + \y k(1+\y) | \\
&= k(1+\y),
\end{align}
where $k < k(1+\y)$ for all $\y \in (0,1)$.

Then we need
\begin{align}
    \frac{k}{1-\y} - k(1+\y) &> C \\
    \Rightarrow k \lp \frac{1}{1-\y} - (1+\y) \rp &> C \\
    \Rightarrow k &> \frac{C}{\frac{1}{1-\y} - (1+\y)} = \frac{C (1-\y)}{\y^2}.
\end{align}

and 
\begin{align}
    k(1+\y) - k &> C \\
    \Rightarrow k \y &> C \\
    \Rightarrow k &> \frac{C}{\y}.
\end{align}

Set $k = \max \lp \frac{C (1-\y)}{\y^2}, \frac{C}{\y} \rp$. 

\hfill $\blacksquare$

\subsection{Theorem 1}

\begin{theorem} \label{appendix:thm1}
{\normalfont \textbf{(Value error as a function of Bellman error).}}
For any state-action pair~$(s,a)\in \mathcal{S} \times \mathcal{A}$, the value error~$\Delta(s,a)$ can be defined as a function of the Bellman error~$\e$ 
\begin{equation}%
    \Delta(s,a) = \frac{1}{1 - \y} \E_{(s',a') \sim d^\pi(\cdot|s,a)} [\e(s',a')].
\end{equation}
\end{theorem}

\textit{Proof.} Our proof follows similar steps to the proof of Lemma 6.1 in \citep{kakade2002approximately} and likely others. 

First by definition:
\begin{align}
    \Delta(s,a) &:= Q(s,a) - Q^\pi(s,a) \\
    \Rightarrow Q^\pi(s,a) &= Q(s,a) - \Delta(s,a). 
\end{align}

Then we can decompose value error:
\begin{align}
    \Delta(s,a) &= Q(s,a) - Q^\pi(s,a) \\
    &= Q(s,a) - (r + \y \E_\pi[Q^\pi(s',a')]) \\
    &= Q(s,a) - (r + \y \E_\pi[Q(s',a') - \Delta(s',a')]) \\
    &= Q(s,a) - (r + \y \E_\pi[Q(s',a')]) + \y \E_\pi[\Delta(s',a')] \\
    &= \e(s,a) + \y \E_\pi[\Delta(s',a')].
\end{align}

By treating $\Delta(s,a)$ as a value function and $\e(s',a')$ as the reward, we can see that:
\begin{equation}
    \Delta(s,a) = \frac{1}{1 - \y} \E_{(s',a') \sim d^\pi(\cdot|s,a)} [\e(s',a')].
\end{equation}

Note that this theorem can also be applied to finite horizon MDPs, by either considering a definition of $d^\pi$ which accounts for the finite horizon, $d^\pi(s',a'|s,a) = \frac{1}{\sum_{t=0}^{T-1} \gamma^t} \sum_{t=0}^{T-1} \y^t p^\pi((s,a) \rightarrow s', t) \pi(a'|s')$, or by transforming the finite horizon MDP into an infinite horizon MDP by considering episode termination to be a terminal state which loops infinitely upon itself. 

\hfill $\blacksquare$

\subsection{Proposition 4}

\begin{proposition}%
{\normalfont \textbf{(Bounds on value error).}}
Let $C_\text{max}=\max_{s,a} |\e(s,a)|$ and $C_\text{avg}=\E_{(s,a) \sim d^\pi} [|\e(s,a)|]$. 
\begin{enumerate}[nosep]
    \item $\frac{C_\text{max}}{1+\y} \leq \max_{s,a} |\Delta(s,a)| \leq \frac{C_\text{max}}{1-\y}$,
    \item $\frac{C_\text{avg}}{1+\y} \leq \E_{(s,a) \sim d^\pi} [|\Delta(s,a)|] \leq \frac{C_\text{avg}}{1-\y}$,
\end{enumerate}
Furthermore, for any MDP and policy, there exists a value function such that the upper bound is an equality, and there exists a MDP, policy, and value function such that the lower bound is an equality. 
\end{proposition}

\textit{Proof.}

Upper bound on $\max_{s,a} |\Delta(s,a)|$: For all $(s,a) \in \mathcal{S} \times \mathcal{A}$:
\begin{align}
    \Delta(s,a) &= \frac{1}{1 - \y} \E_{(s',a') \sim d^\pi(\cdot|s,a)} [\e(s',a')] \\
    \Rightarrow |\Delta(s,a)| &= | \frac{1}{1 - \y} \E_{(s',a') \sim d^\pi(\cdot|s,a)} [\e(s',a')] | \\
    &\leq \frac{1}{1 - \y} \E_{(s',a') \sim d^\pi(\cdot|s,a)} [|\e(s',a')|] \\
    &\leq \frac{1}{1 - \y}  \max_{(s,a) \in \mathcal{S} \times \mathcal{A}} |\e(s,a)|\\
    \Rightarrow \max_{(s,a)} |\Delta(s,a)| &\leq \frac{1}{1 - \y}  \max_{(s,a) \in \mathcal{S} \times \mathcal{A}} |\e(s,a)|.
\end{align}
A proof of a similar relationship can also be found in \citet{bertsekas1995dynamic},~(Proposition 6.1, page 262).

Lower bound on $\max_{s,a} |\Delta(s,a)|$: For all $(s,a) \in \mathcal{S} \times \mathcal{A}$:
\begin{align}
    \e(s,a) &= \Delta(s,a) - \y \E_{s', \pi}[ \Delta(s',a') ] \\
    \Rightarrow \Delta(s,a) &= \e(s,a) + \y \E[\Delta(s',a')] \\
    \Rightarrow |\Delta(s,a)| &= | \e(s,a) + \y \E[\Delta(s',a')] | \\
    \Rightarrow \max_{(s,a)} |\Delta(s,a)| &= \max_{(s,a)} | \e(s,a) + \y \E[\Delta(s',a')] | \\
    &\geq \max_{(s,a)} |\e(s,a)| - \y | \E[\Delta(s',a')] | \\
    &\geq \max_{(s,a)} |\e(s,a)| - \y \max_{(s,a)}  | \E[\Delta(s',a')] | \\
    &\geq \max_{(s,a)} |\e(s,a)| - \y \max_{(s,a)} |\Delta(s,a)|\\
    \Rightarrow (1 + \y) \max_{(s,a)} |\Delta(s,a)| &\geq \max_{(s,a)} |\e(s,a)| \\
    \Rightarrow \max_{(s,a)} |\Delta(s,a)| &\geq \frac{1}{1 + \y} \max_{(s,a)} |\e(s,a)|. \\
\end{align}

Upper bound on $\E_{(s,a) \sim d^\pi} [|\Delta(s,a)|]$: For all $(s,a) \in \mathcal{S} \times \mathcal{A}$:
\begin{align}
    \Delta(s,a) &= \frac{1}{1 - \y} \E_{(s',a') \sim d^\pi(\cdot|s,a)} [\e(s',a')] \\
    \Rightarrow |\Delta(s,a)| &= \frac{1}{1 - \y} |\E_{(s',a') \sim d^\pi(\cdot|s,a)} [\e(s',a')]| \\
    \Rightarrow \E_{d^\pi} [ |\Delta(s,a)| ] &= \frac{1}{1 - \y} \E_{d^\pi} [ |\E_{(s',a') \sim d^\pi(\cdot|s,a)} [\e(s',a')]| ] \\
    &\leq \frac{1}{1 - \y} \E_{d^\pi} [ \E_{(s',a') \sim d^\pi(\cdot|s,a)} [|\e(s',a')|] ] \\
    &= \frac{1}{1 - \y} \E_{d^\pi} [|\e(s',a')|].
\end{align}

Lower bound on $\E_{(s,a) \sim d^\pi} [|\Delta(s,a)|]$: For all $(s,a) \in \mathcal{S} \times \mathcal{A}$:
\begin{align}
    \e(s,a) &= \Delta(s,a) - \y \E_{s', \pi}[ \Delta(s',a') ] \\
    \Rightarrow \Delta(s,a) &= \e(s,a) + \y \E[\Delta(s',a')] \\
    \Rightarrow |\Delta(s,a)| &= | \e(s,a) + \y \E[\Delta(s',a')] | \\
    \Rightarrow \E_{d^\pi} [  |\Delta(s,a)| ] &= \E_{d^\pi} [ | \e(s,a) + \y \E[\Delta(s',a')] | ] \\
    &\geq \E_{d^\pi} [ | \e(s,a) | - \y | \E[\Delta(s',a')] | ] \\
    &= \E_{d^\pi} [ | \e(s,a) | ] - \y \E_{d^\pi} [ | \E[\Delta(s',a')] | ] \\
    \Rightarrow (1 + \y ) \E_{d^\pi} [  |\Delta(s,a)| ] &\geq \E_{d^\pi} [ | \e(s,a) | ] \\
    \Rightarrow \E_{d^\pi} [  |\Delta(s,a)| ] &\geq \frac{1}{1 + \y} \E_{d^\pi} [ | \e(s,a) | ].
\end{align}

Equality on upper bound: For any MDP, policy $\pi$, and constant $C$, let $Q(s,a) = Q^\pi + \frac{C}{1 - \y}$ for all $(s,a) \in \mathcal{S} \times \mathcal{A}$. It follows that $\e(s,a) = C$ and $\Delta(s,a) = \frac{C}{1 - \y}$ for all $(s,a) \in \mathcal{S} \times \mathcal{A}$. 

Equality on lower bound: Consider a two-state, single action MDP $\mathcal{S}=\{ s_1,s_2 \}, \mathcal{A}=\{ a \}$ such that $(s_1,a) \rightarrow s_2$ and $(s_2,a) \rightarrow s_1$. Let $Q(s_1,a) = Q^\pi + \frac{C}{1 + \y}$ and $Q(s_2,a) = Q^\pi - \frac{C}{1 + \y}$ for some constant $C$. It follows that $|\e(s,a)| = C$ and $|\Delta(s,a)| = \frac{C}{1 + \y}$ for all $(s,a) \in \mathcal{S} \times \mathcal{A}$. 

\hfill $\blacksquare$

\subsection{Corollary 1} %

\begin{corollary} %
{\normalfont \textbf{(The Bellman equation is not unique over incomplete datasets).}}
If there exists a state-action pair $(s',a')$ not contained in the dataset $\mathcal{D}$, where the state-action occupancy $d^\pi(s',a'|s,a) > 0$ for some $(s,a) \in \mathcal{D}$, then there exists a value function and $C>0$ such that 
\begin{enumerate}[nosep]
    \item For all $(\hat s, \hat a) \in \mathcal{D}$, the Bellman error $\e(\hat s,\hat a) = 0$.
    \item There exists $(s,a) \in \mathcal{D}$, such that the value error $\Delta(s,a) = C$.
\end{enumerate}
\end{corollary}

\textit{Proof.} This is a direct consequence of \autoref{appendix:thm1}. Let $\mathcal{D}'$ contain the set of state-action pairs $(s',a')$ not contained in the dataset $\mathcal{D}$, where the state-action occupancy $d^\pi(s',a'|s,a) > 0$. Let $\mathcal{D}_\text{unique}$ be the set of unique state-action pairs in $\mathcal{D}$. It follows that
\begin{align}
    \Delta(s,a) &= \frac{1}{1 - \y} \E_{(s',a') \sim d^\pi(\cdot|s,a)} [\e(s',a')] \\
    &= \frac{1}{1 - \y} \sum_{(s',a') \sim \mathcal{D}_\text{unique}} d^\pi(s',a'|s,a) \e(s',a') + \frac{1}{1 - \y} \sum_{(s',a') \sim \mathcal{D}'} d^\pi(s',a'|s,a) \e(s',a'). 
\end{align}

Recall
\begin{equation}
    \e(s,a) := Q(s,a) - \E_{r,s' \sim p,a' \sim \pi} \lb r + \y Q(s',a') \rb,
\end{equation}
and there exists at least one $Q(s,a)$, such that $(s,a) \in \mathcal{D}'$. Since the sets $\mathcal{D}$ and $\mathcal{D}'$ are distinct, it follows that there exists a function $Q$ such that $\e(s,a) = 0$ for all $(s,a) \in \mathcal{D}$, 
but $\frac{1}{1 - \y} \sum_{(s',a') \sim \mathcal{D}'} d^\pi(s',a'|s,a) \e(s',a') = C$.

\hfill $\blacksquare$

\subsection{Proposition 5}

\begin{proposition}%
{\normalfont \textbf{(FQE improvement condition).}} Let $Q_1, Q_2$ be value functions, where $Q_2 = \mathcal{T} Q_1$. 
If $\y \cdot \max_{(s,a) \in \mathcal{D}} |\E_{(s',a') \sim p(\cdot|s,a), \pi}[\Delta_{Q_1}(s',a')]| < \max_{(s,a) \in \mathcal{D}} |\Delta_{Q_1}(s,a)|$ then $\max_{(s,a) \in \mathcal{D}} |\Delta_{Q_2}(s,a)| < \max_{(s,a) \in \mathcal{D}} |\Delta_{Q_1}(s,a)|$.
\end{proposition}

\textit{Proof.} 

For all $(s,a) \in \mathcal{D}$: 
\begin{align}
    \Delta_2(s,a) &= \mathcal{T} Q_1(s,a) - Q^\pi(s,a) \\
    &= \mathcal{T} Q_1(s,a) - \mathcal{T} Q^\pi(s,a) \\
    &= \E_{(s',a') \sim p(\cdot|s,a), \pi}[\y (Q_1(s',a') - Q^\pi(s',a'))] \\
    &= \E_{(s',a') \sim p(\cdot|s,a), \pi}[\y \Delta_1(s',a')] \\
\Rightarrow |\Delta_2(s,a)| &= \y |\E_{(s',a') \sim p(\cdot|s,a), \pi}[\Delta_1(s',a')]|. 
\end{align}

It follows that for all $(s,a) \in \mathcal{D}$ if
\begin{align}
    \y \cdot \max_{(s,a) \in \mathcal{D}} |\E_{(s',a') \sim p(\cdot|s,a), \pi}[\Delta_1(s',a')]| &< \max_{(s,a) \in \mathcal{D}} |\Delta_1(s,a)| \\
    \Rightarrow \max_{(s,a) \in \mathcal{D}} |\Delta_2(s,a)| &< \max_{(s,a) \in \mathcal{D}} |\Delta_1(s,a)|. 
\end{align}

\hfill $\blacksquare$

\clearpage
\section{Learning Curves}

\subsection{BRM}

\begin{figure}[ht]
    \centering
    \begin{subfigure}{\textwidth}
    \includegraphics[width=\textwidth]{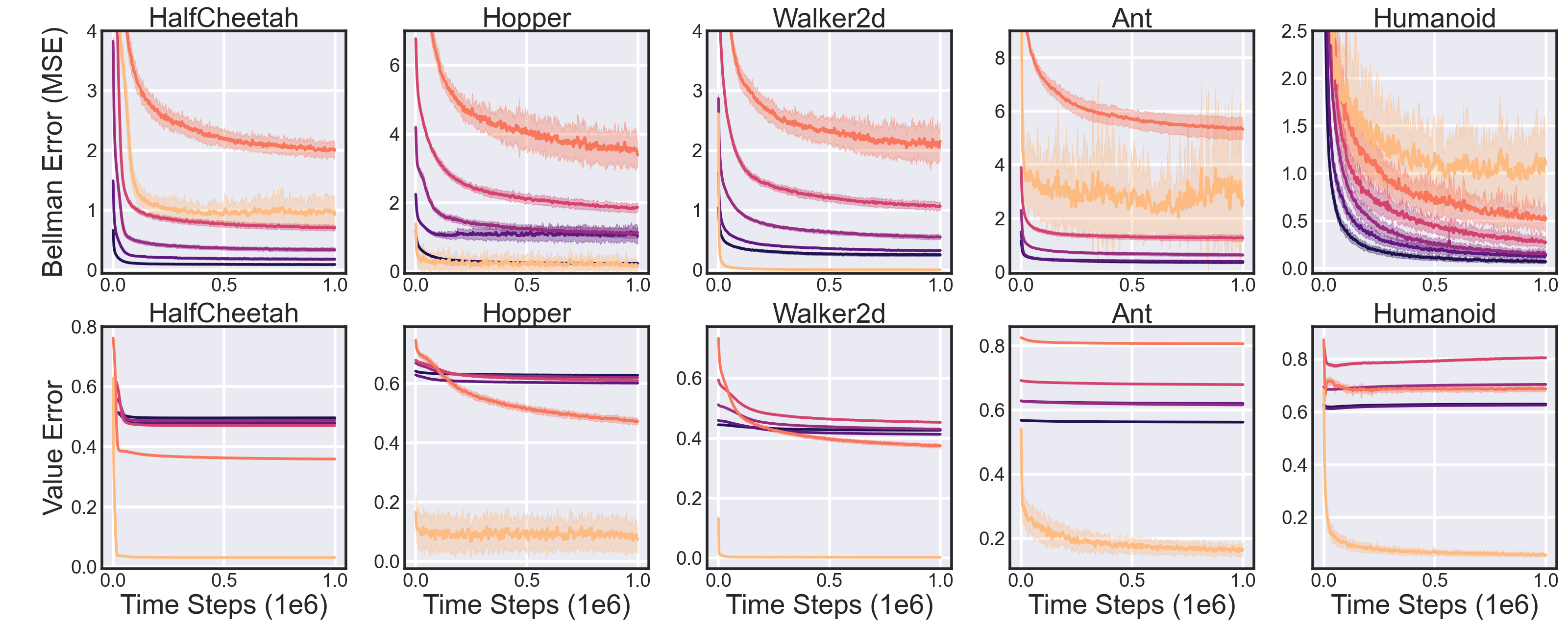}
    \caption{Evaluated with the Training Dataset}\label{fig:residual_training}
    \end{subfigure}
    
    \vspace{0.5em}
    \begin{subfigure}{\textwidth}
    \includegraphics[width=\textwidth]{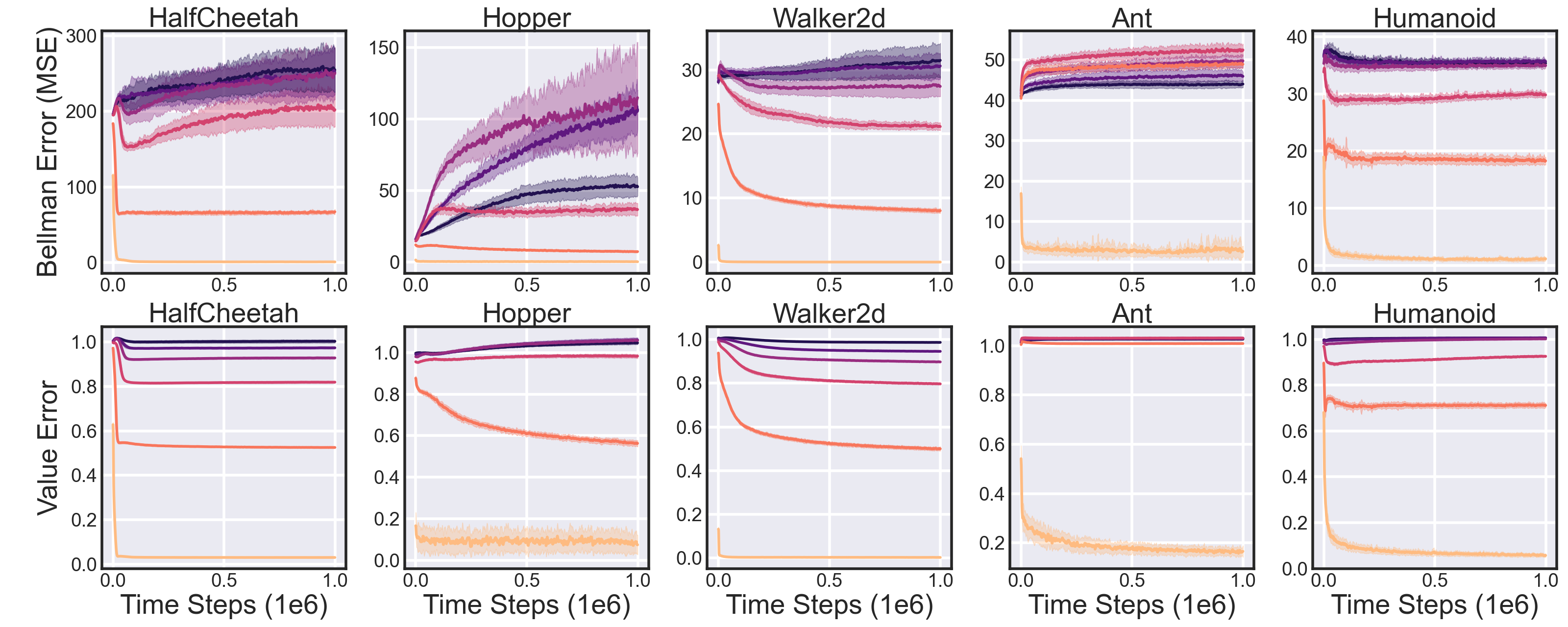}
    \caption{Evaluated with the On-Policy Dataset}
    \end{subfigure}

    \vspace{0.5em}
    \small Noise level: \qquad \cblock{my_warm_1} On-policy ($0.0$) \qquad \cblock{my_warm_2} $0.1$ \qquad \cblock{my_warm_3} $0.2$ \qquad \cblock{my_warm_4} $0.3$ \qquad \cblock{my_warm_5} $0.4$ \qquad \cblock{my_warm_6} $0.5$
    \caption{Visualizing the learning curves of the Bellman error and value error of value functions trained by BRM evaluated over the training dataset (used in \autoref{table:OPE_obj}) and on-policy dataset (used in \autoref{table:correlation}). The shaded area captures the standard deviation over 10 seeds.} \label{fig:residual} %
\end{figure}

\clearpage
\subsection{FQE} \label{appendix:FQEcurves}

\begin{figure}[ht]
    \centering
    \begin{subfigure}{\textwidth}
    \includegraphics[width=\textwidth]{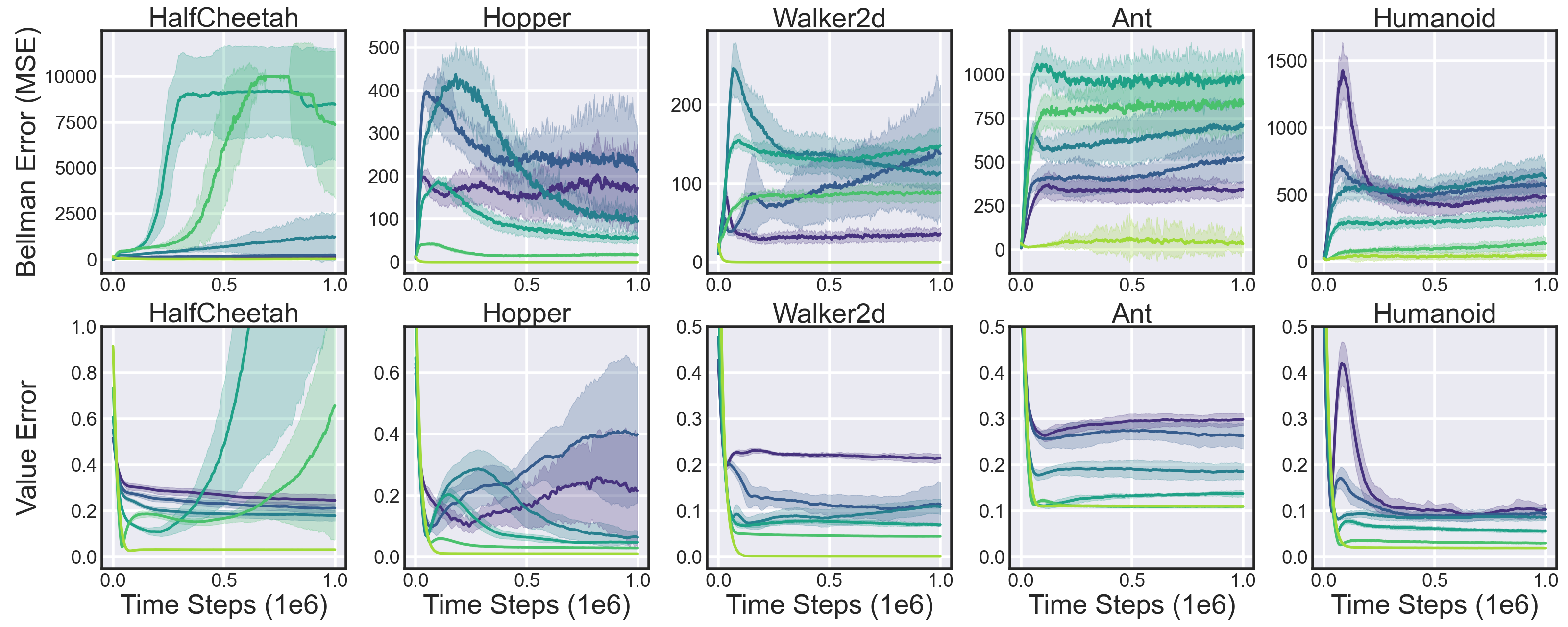}
    \caption{Evaluated with the Training Dataset}
    \end{subfigure}
    
    \vspace{0.5em}
    \begin{subfigure}{\textwidth}
    \includegraphics[width=\textwidth]{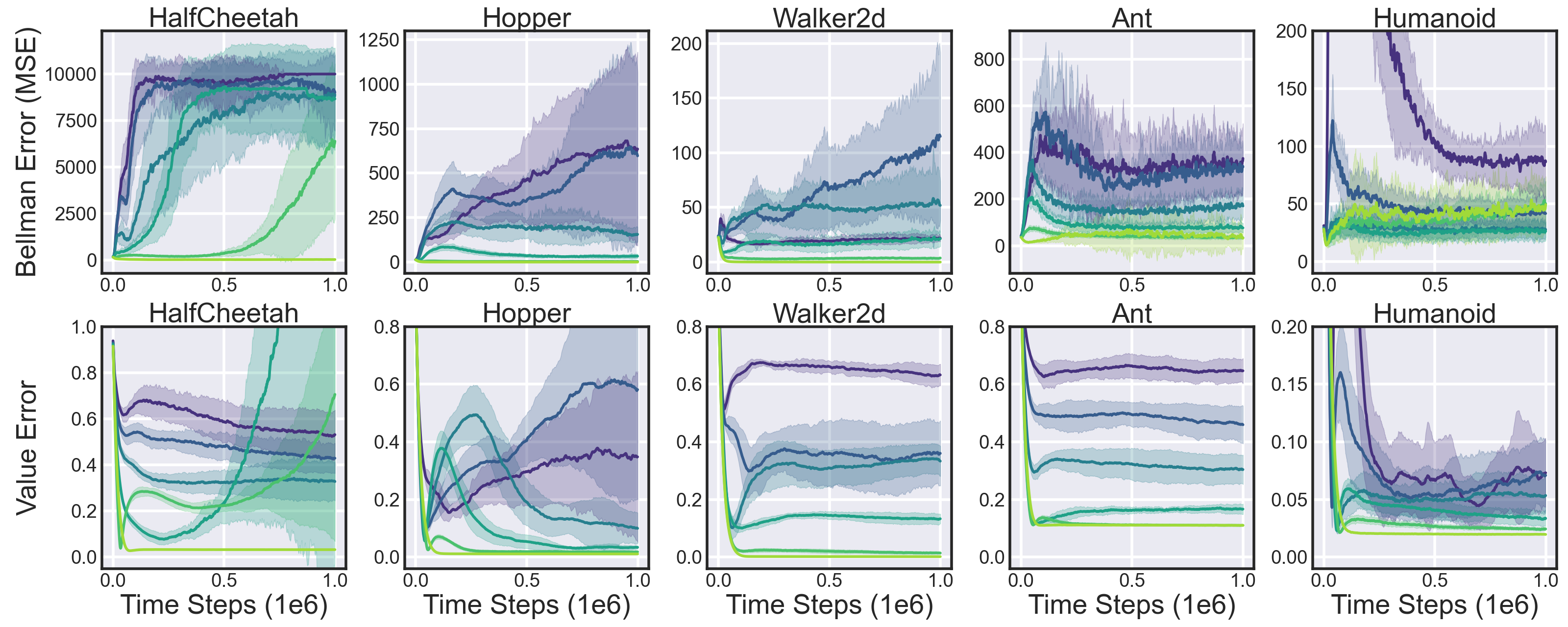}
    \caption{Evaluated with the On-Policy Dataset}
    \end{subfigure}

    \vspace{0.5em}
    \small Noise level: \qquad \cblock{my_cold_1} On-policy ($0.0$) \qquad \cblock{my_cold_2} $0.1$ \qquad \cblock{my_cold_3} $0.2$ \qquad \cblock{my_cold_4} $0.3$ \qquad \cblock{my_cold_5} $0.4$ \qquad \cblock{my_cold_6} $0.5$    
    \caption{Visualizing the learning curves of the Bellman error and value error of value functions trained by FQE evaluated over the training dataset (used in \autoref{table:OPE_obj}) and on-policy dataset (used in \autoref{table:correlation}). The shaded area captures the standard deviation over 10 seeds. The Bellman error of individual trials is clipped to 10k for visual clarity (only relevant for HalfCheetah).} \label{fig:fqe} %
\end{figure}

\clearpage
\section{What if...}\label{appendix:sec:whatif}

\subsection{...we compare the absolute Bellman error to absolute value error?}\label{appendix:absabs}

\autoref{fig:both} compares the mean squared Bellman error to the absolute value error. In \autoref{appendix:fig:absabs} we compare the absolute Bellman error with the absolute value error, and in \autoref{appendix:fig:powpow} we compare the mean squared Bellman error with the mean squared value error. We find that our observations regarding the relative ordering of the Bellman error and value error are unchanged. %
\begin{figure}[ht]
    \centering
    \includegraphics[width=\textwidth]{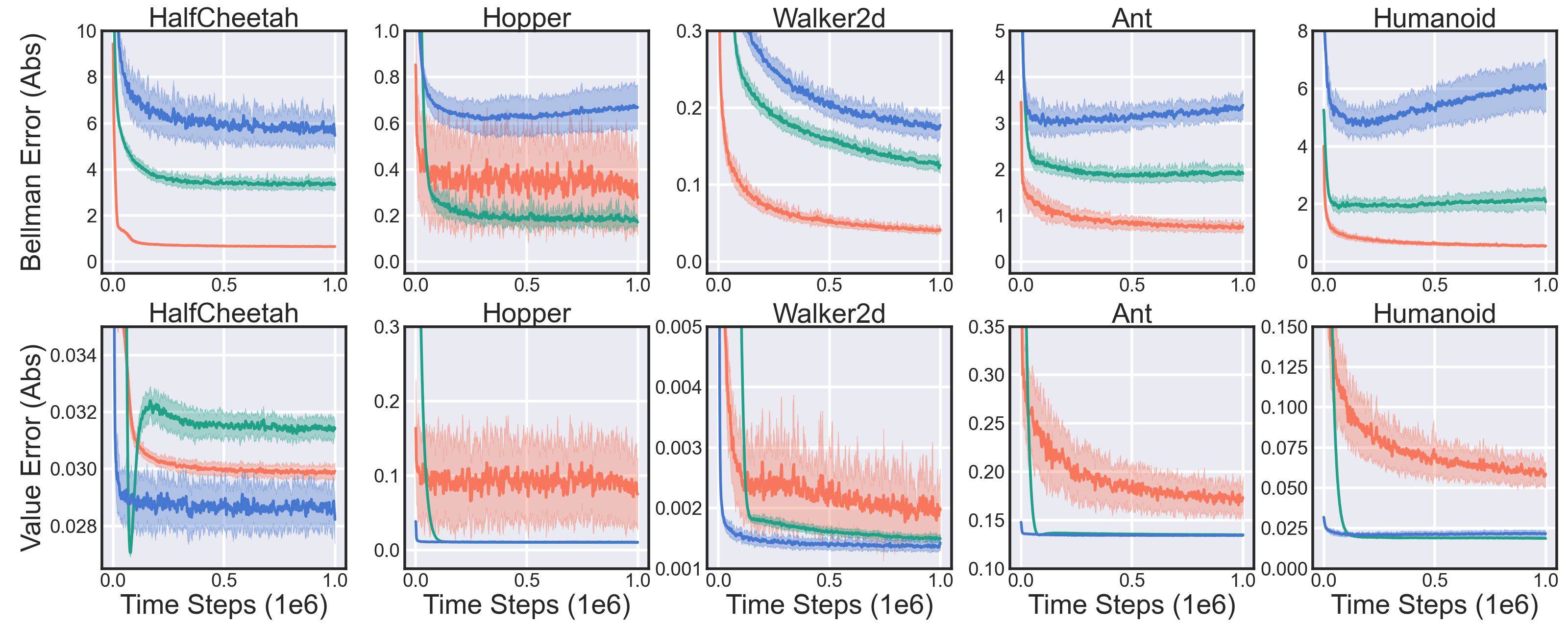}
    \small \cblock{my_warm_2} BRM \qquad \cblock{my_cold_3} FQE \qquad \cblock{my_blue} MC
\caption[]{Comparing the absolute Bellman error with the absolute value error. The shaded area captures the standard deviation over 10 seeds. Both algorithms are trained using on-policy data collected by the target policy.} \label{appendix:fig:absabs}
\end{figure}

\begin{figure}[ht]
    \centering
    \includegraphics[width=\textwidth]{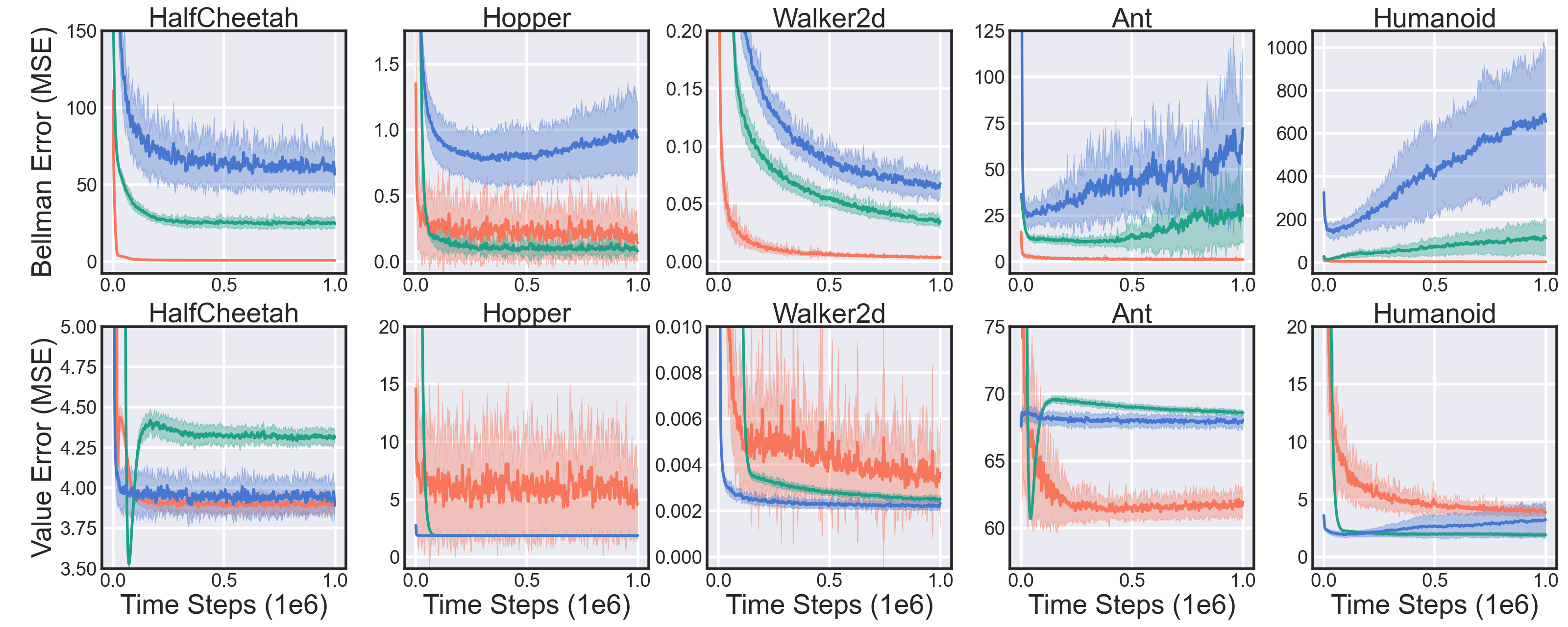}
    \small \cblock{my_warm_2} BRM \qquad \cblock{my_cold_3} FQE \qquad \cblock{my_blue} MC
\caption[]{Comparing the mean squared Bellman error with the mean squared value error. The shaded area captures the standard deviation over 10 seeds. Both algorithms are trained using on-policy data collected by the target policy.} \label{appendix:fig:powpow}
\end{figure}

\clearpage
\subsection{...we use less data?}

\autoref{fig:both} uses datasets of 1M transitions. 
In \autoref{appendix:fig:50k} we repeat the experiment in \autoref{fig:both} with a dataset of 50k transitions. We find that our observations regarding the relative ordering of the Bellman error and value error are unchanged.

\begin{figure}[ht]
    \centering
    \includegraphics[width=\textwidth]{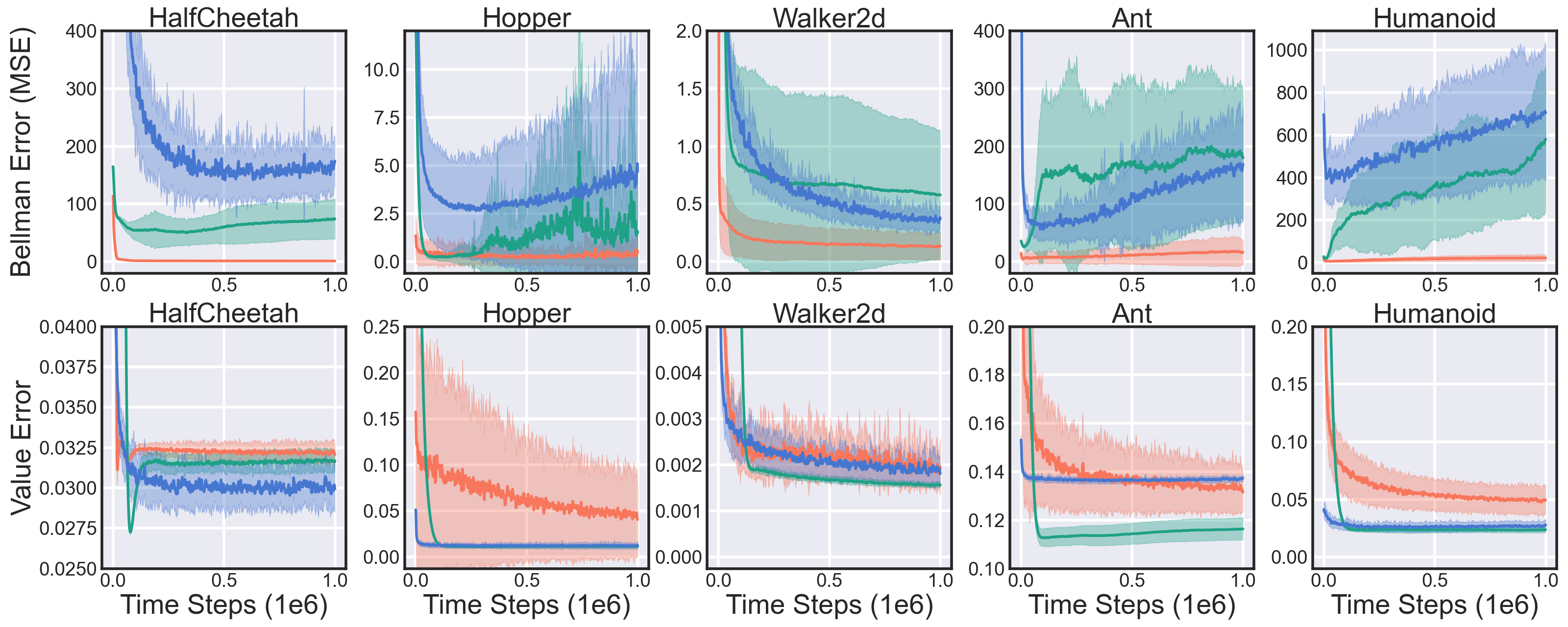}
    \small \cblock{my_warm_2} BRM \qquad \cblock{my_cold_3} FQE \qquad \cblock{my_blue} MC
\caption[]{Comparing the mean squared Bellman error with the absolute value error, using a dataset of 50k. The shaded area captures the standard deviation over 10 seeds. Both algorithms are trained using on-policy data collected by the target policy.} \label{appendix:fig:50k}
\end{figure}

\subsection{...we use training error instead of test error?}

\begin{figure}[ht]
    \centering
    \includegraphics[width=\textwidth]{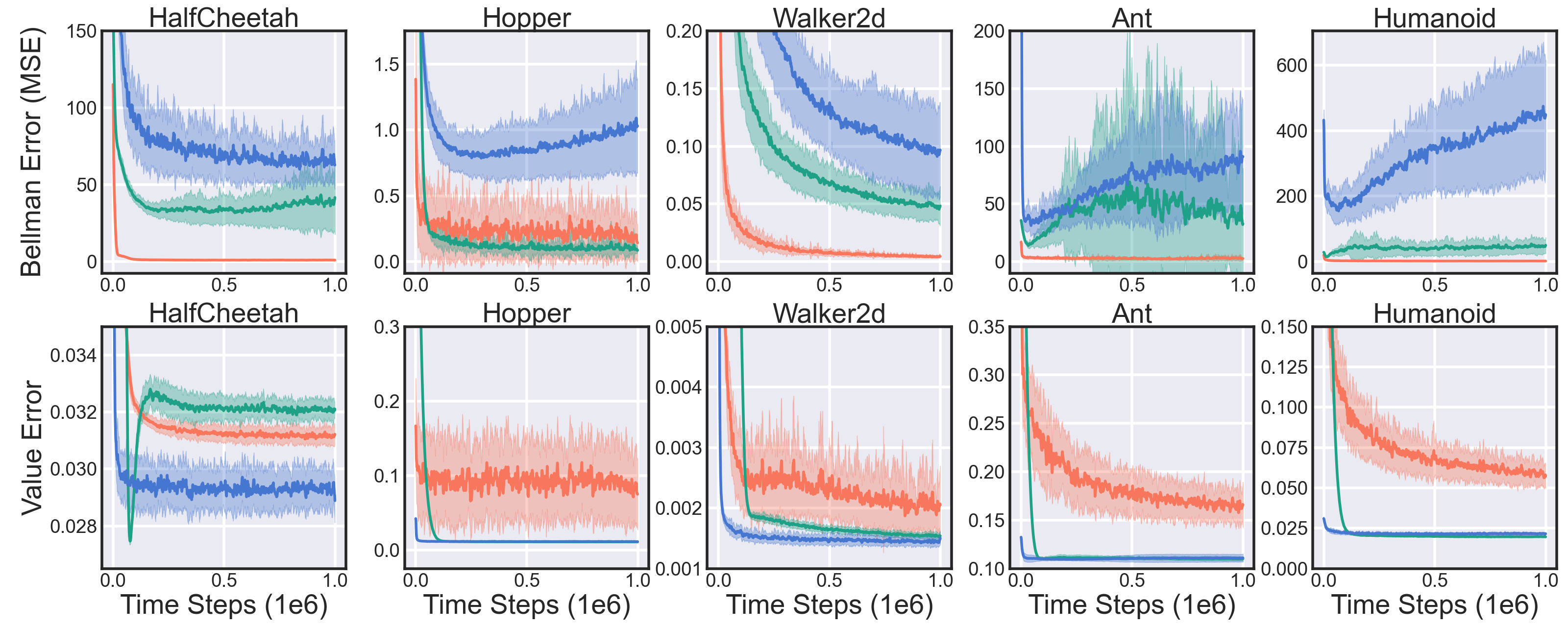}
    \small \cblock{my_warm_2} BRM \qquad \cblock{my_cold_3} FQE \qquad \cblock{my_blue} MC
\caption[]{Comparing the mean squared Bellman error with the absolute value error, evaluated on the training dataset, rather than a held-out test set. The shaded area captures the standard deviation over 10 seeds. Both algorithms are trained using on-policy data collected by the target policy.} \label{fig:training_test}
\end{figure}

\autoref{fig:both} evaluates the Bellman error and the value error on a held-out test set. In \autoref{fig:training_test} we evaluate the error terms on the training dataset, rather than the held-out test set. We find that our observations regarding the relative ordering of the Bellman error and value error are unchanged.

\clearpage
\subsection{...we use train for longer?}

\autoref{fig:both} displays the learning curves for 1M time steps. 
In \autoref{appendix:fig:10m} we train for 10M time steps (using the same dataset). We find that our observations regarding the relative ordering of the Bellman error and value error are unchanged.

\begin{figure}[ht]
    \centering
    \includegraphics[width=\textwidth]{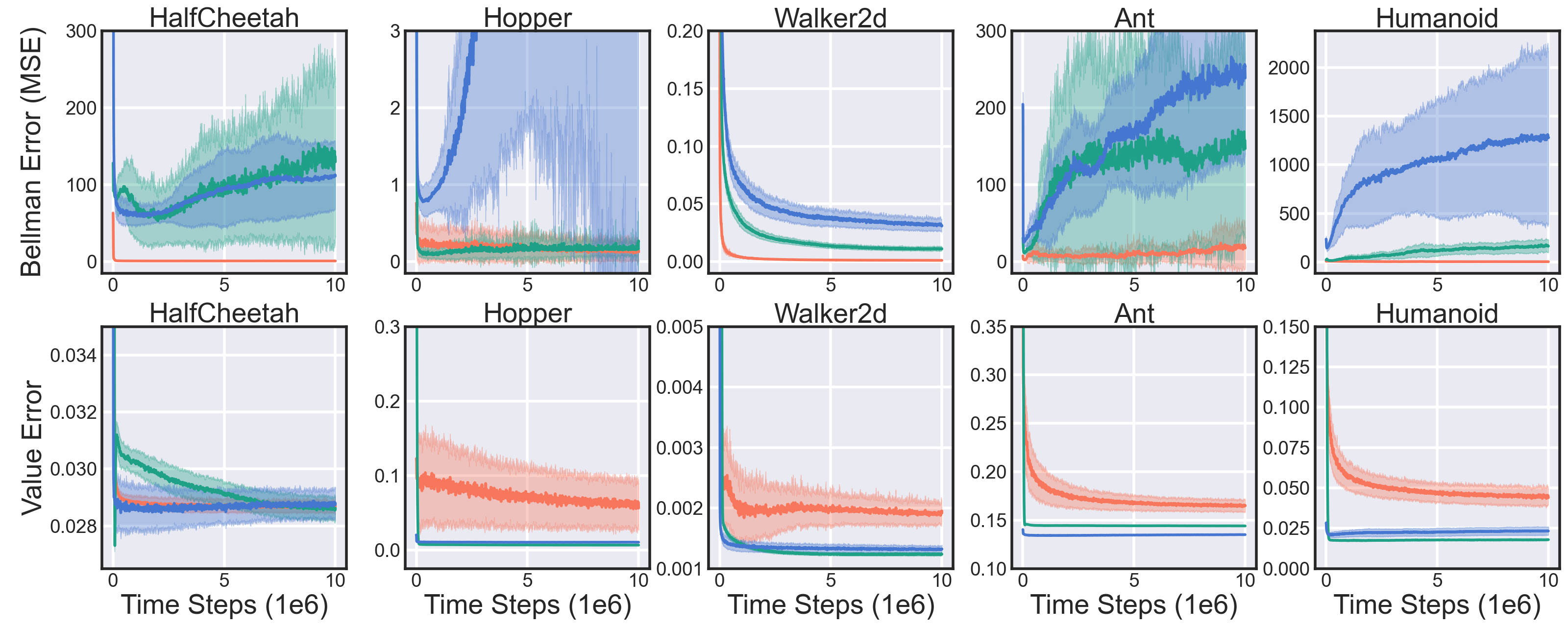}
    \small \cblock{my_warm_2} BRM \qquad \cblock{my_cold_3} FQE \qquad \cblock{my_blue} MC
\caption[]{Comparing the mean squared Bellman error with the absolute value error while training for 10M time steps (the dataset of 1M is unchanged from \autoref{fig:both}). The shaded area captures the standard deviation over 10 seeds. Both algorithms are trained using on-policy data collected by the target policy.} \label{appendix:fig:10m}
\end{figure}

\clearpage
\section{Experiment Details} \label{appendix:details}

\textbf{Software.} Software versions used were as follows:
\begin{itemize}[nosep]
    \item Python 3.6
    \item Pytorch 1.8.0~\citep{paszke2019pytorch}
    \item Gym 0.17.0~\citep{OpenAIGym}
    \item MuJoCo 1.50\footnote{License information: \url{https://www.roboti.us/license.html}}~\citep{mujoco}%
    \item mujoco-py 1.50.1.1
\end{itemize}
-v3 versions of the MuJoCo environments were used.

\textbf{Hyperparameters.} FQE and BRM used the same hyperparameters and architecture, as described in \autoref{appendix:table:hyperparameters}. These hyperparameters were chosen to match TD3 and SAC~\citep{fujimoto2018addressing, haarnoja2018soft}, state of the art off-policy RL algorithms used in the MuJoCo domain. Following these algorithms, both FQE and BRM set the discount factor $\y$ to $0$ for terminal states (and use $0.99$ otherwise). 
FQE uses Polyak averaging for the target network update. Given parameters $\ta$ of the current network, the parameters of the target network $\bar \ta$ are updated by the following after each time step:
\begin{equation}
    \bar \ta \leftarrow (1 - \tau) \bar \ta + \tau \ta,
\end{equation}
where $\tau$ is the target update rate. This rule is a commonly-used update rule by many off-policy RL algorithms for continuous actions~\citep{DDPG, fujimoto2018addressing, haarnoja2018applications}. 

\begin{table}[ht]
\centering
\begin{tabular}{cll}
\toprule
& Hyperparameter & Value \\
\midrule
\multirow{5}{*}{\shortstack{Network\\Hyperparameters}} & Optimizer & Adam~\citep{adam} \\
& Learning rate & 3e-4 \\
& Mini-batch size      & 256 \\
& Target update rate (FQE)   & 5e-3 \\
& Discount factor      & 0.99 \\
& Terminal Discount factor      & 0.0 \\
\midrule
\multirow{3}{*}{Architecture} & Network Hidden dim & 256 \\
& Network Hidden layers & 2 \\
& Activation function & ReLU \\
\bottomrule
\end{tabular}
\caption{Hyperparameters and architecture.} \label{appendix:table:hyperparameters}
\end{table}

\textbf{Target policy.} In each experiment, we evaluate the discounted return of a deterministic target policy taken from TD3~\citep{fujimoto2018addressing} trained for 3 million time steps. Our implementation of TD3 is based directly off of the author-provided Github (\url{https://github.com/sfujim/TD3}). For all experiments, the discounted return uses a discount factor of $\y=0.99$.

\textbf{Dataset and behavior policy.} Datasets are collected by using a noisy variation of the target policy $\pi_t$. For a noise level $n \in [0.0,0.1,0.2,0.3,0.4,0.5]$, the behavior policy $\pi_b$ is defined as:
\begin{equation}
    \pi_b(s) = \begin{cases}
    \pi_t(s) + \N(0,n), & \text{with }p=1-n,\\
    \text{uniform random action} & \text{with }p=n.
    \end{cases}
\end{equation}
Unless stated otherwise, experiments use a dataset of 1 million transitions, matching the replay buffer size of TD3/SAC.

\textbf{Metrics and evaluation datasets.} The main metrics used are the Bellman error and the value error. Given an evaluation dataset $\mathcal{D}_e$ and value function $Q$, the mean-squared Bellman error is computed by:
\begin{equation}
\frac{1}{|\mathcal{D}_e|} \sum_{(s,a,r,s') \sim \mathcal{D}_e, a' \sim \pi} \lp Q(s,a) - (r + \y Q(s',a')) \rp^2.    
\end{equation}
Given an evaluation dataset $\mathcal{D}_e$ and value function $Q$, the normalized absolute value error is computed by:
\begin{equation}
\frac{1}{K |\mathcal{D}_e|} \sum_{(s,a,r,s') \sim \mathcal{D}_e, a' \sim \pi} |Q(s,a) - Q^\pi(s,a)|.
\end{equation}
$Q^\pi(s,a)$ is computed near exactly by resetting the MuJoCo environment to the specific state-action pair $(s,a)$ and running the policy for $1000$ time steps (the environment and policy are deterministic, so one trajectory is sufficient to estimate the true value). Value error is normalized by a per-environment constant equal to the average true value $Q^\pi$ over the on-policy evaluation dataset $\mathcal{D}_{\text{on-policy}}$:
\begin{equation}
    K = \frac{1}{|\mathcal{D}_{\text{on-policy}}|} \sum_{(s,a,r,s') \sim \mathcal{D}_{\text{on-policy}}} Q^\pi(s,a).
\end{equation}
We report the values of $K$ used in \autoref{appendix:table:K}.
 
\begin{table}[ht]
\centering
\begin{tabular}{lc}
\toprule
Environment & $K$ \\
\midrule
HalfCheetah & 1382.35 \\
Hopper      & 388.56  \\
Walker2d    & 529.12  \\
Ant         & 580.90  \\
Humanoid    & 571.29  \\
\bottomrule
\end{tabular}
\caption{Values of the per-environment normalizing constant $K$, used to normalize value error for better interpretability across tasks.} \label{appendix:table:K}
\end{table}

Evaluation datasets are each collected by using the same set of behavior policies used to generate the training datasets, in other words with noise levels $[0.0, 0.1, 0.2, 0.3, 0.4, 0.5]$. Each evaluation dataset contains 1000 transitions, and is gathered by collecting 50k transitions, and then selecting 1000 of the 50k transitions with uniform probability. Error terms are computed over an evaluation dataset of 1000 transitions, generated in similar fashion as the training datasets. 

\textbf{Termination.} The MuJoCo environments are time-limited at $1000$ time steps. Since using time-limited termination (which is not represented in the state) breaks the Markov property, we follow standard practice~\citep{fujimoto2018addressing, haarnoja2018soft} and only consider a state terminal if termination occurs before $1000$ time steps. In other words, $\y=0$ if a state $s_t$ is both terminal and $t < 1000$, and $\y=0.99$ otherwise. 

\textbf{Outliers.} In the experiments for \autoref{table:OPE_obj}, for HalfCheetah with the $0.1$ and $0.2$ datasets the value function trained by FQE diverged on several seeds (see \autoref{appendix:FQEcurves} for visualization), these trials were removed when computing the values in the table. Note that this does not affect the conclusions made, as we are interested in the existence of problems with the Bellman error, not comparing the performance of BRM and FQE. 

Tables (\ref{table:correlation} \& \ref{table:correlation_FQE}) report Pearson's correlation coefficient. Since this measure is not robust to outliers, for FQE we remove the 30\% of data points with the highest Bellman error terms (functions trained with BRM had no obvious outliers).

\textbf{Single trajectory experiments.} In \autoref{fig:one_traj} and \autoref{appendix:one_traj} experiments were performed over a single on-policy trajectory. This trajectory was collected by the expert target policy. Unlike the other experiments, termination was used so that the dataset was complete (contained every relevant transition). To reduce the importance of propagating values, we added the average reward in the dataset, scaled by the horizon (i.e.\ $\frac{1}{1-\y}=100$), to the terminal reward $r_{T-1} \leftarrow r_{T-1} + \frac{\y}{(1 - \y) T}\sum_{i=0}^{T-1} r_i$, where $T=1000$ is the total numbers of transitions in the trajectory. Additionally, algorithms were trained for 250k time steps rather than 1M. 

\end{document}